\documentclass[manuscript]{acmart}

\usepackage{acronym}
\usepackage[linesnumbered]{algorithm2e}
\usepackage{bbold}
\usepackage{tabularx}
\usepackage{multirow}
\usepackage[inline]{enumitem}
\usepackage{subcaption}

\AtBeginDocument{%
  \providecommand\BibTeX{{%
    \normalfont B\kern-0.5em{\scshape i\kern-0.25em b}\kern-0.8em\TeX}}}

\setcopyright{acmlicensed}
\acmJournal{TIIS}
\acmYear{2023} \acmVolume{1} \acmNumber{1} \acmArticle{1} \acmMonth{1} \acmPrice{15.00}\acmDOI{10.1145/3625240}

\newcommand{\ie}{i.e.}
\newcommand{\eg}{e.g.}
\newcommand*{\name}[1]{\textsc{#1}}

\usepackage[normalem]{ulem}
\usepackage{color}
\definecolor{Green}{RGB}{0, 200, 0}

\newacro{API}[API]{application programming interface}
\newacro{AI}[AI]{artificial intelligence}
\newacro{AutoML}[AutoML]{automated machine learning}
\newacro{CPC}[CPC]{conditional parallel coordinates}
\newacro{DAG}[DAG]{directed acyclic graph}
\newacro{FATE}[FATE]{fairness, accountability, transparency, and ethics}
\newacro{HPO}[HPO]{hyperparameter optimization}
\newacro{ICE}[ICE]{independent conditional expectation}
\newacro{EMG}[EMG]{electromyography}
\newacro{MCTS}[MCTS]{Monte Carlo tree search}
\newacro{MDS}[MDS]{multidimensional scaling}
\newacro{ML}[ML]{machine learning}
\newacro{PCA}[PCA]{principal component analysis}
\newacro{PDP}[PDP]{partial dependence plot}
\newacro{ROC}[ROC]{receiver operating characteristic}
\newacro{SUS}[SUS]{system usability score}
\newacro{SVM}[SVM]{support-vector machine}
\newacro{XAI}[XAI]{explainable artificial intelligence}

\begin{document}

\title{XAutoML: A Visual Analytics Tool for Understanding and Validating Automated Machine Learning}

\author{Marc-Andr\'e Z\"oller}
\authornote{Corresponding Author}
\email{marc.zoeller@usu.com}
\affiliation{%
  \institution{USU Software AG}
  \streetaddress{R\"uppurrer Str. 1}
  \city{Karlsruhe}
  \state{Baden-Württemberg}
  \country{Germany}
  \postcode{76137}
}

\author{Waldemar Titov}
\email{waldemar.titov@h-ka.de}
\affiliation{%
  \institution{Institute of Ubiquitous Mobility Systems, Karlsruhe University of Applied Sciences}
  \streetaddress{Moltkestr. 30}
  \city{Karlsruhe}
  \state{Baden-Württemberg}
  \country{Germany}
  \postcode{76133}
}

\author{Thomas Schlegel}
\email{thomas.schlegel@h-ka.de}
\affiliation{%
  \institution{Institute of Ubiquitous Mobility Systems, Karlsruhe University of Applied Sciences}
  \streetaddress{Moltkestr. 30}
  \city{Karlsruhe}
  \state{Baden-Württemberg}
  \country{Germany}
  \postcode{76133}
}

\author{Marco F. Huber}
\email{marco.huber@ieee.org}
\affiliation{%
  \institution{Institute of Industrial Manufacturing and Management IFF, University of Stuttgart, and Department Cyber Cognitive Intelligence (CCI), Fraunhofer IPA}
  \streetaddress{Nobelstr. 12}
  \city{Stuttgart}
  \state{Baden-Württemberg}
  \country{Germany}
  \postcode{70569}
}


\begin{abstract}
In the last 10 years, various \ac{AutoML} systems have been proposed to build end-to-end \ac{ML} pipelines with minimal human interaction. Even though such automatically synthesized \ac{ML} pipelines are able to achieve competitive performance, recent studies have shown that users do not trust models constructed by \ac{AutoML} due to missing transparency of \ac{AutoML} systems and missing explanations for the constructed \ac{ML} pipelines. In a requirements analysis study with 36 domain experts, data scientists, and AutoML researchers from different professions with vastly different expertise in \ac{ML}, we collect detailed informational needs for \ac{AutoML}. We propose \name{XAutoML}, an interactive visual analytics tool for explaining arbitrary \ac{AutoML} optimization procedures and \ac{ML} pipelines constructed by \ac{AutoML}. \name{XAutoML} combines interactive visualizations with established techniques from \ac{XAI} to make the complete \ac{AutoML} procedure transparent and explainable. By integrating \name{XAutoML} with \name{JupyterLab}, experienced users can extend the visual analytics with ad-hoc visualizations based on information extracted from \name{XAutoML}. We validate our approach in a user study with the same diverse user group from the requirements analysis. All participants were able to extract useful information from \name{XAutoML}, leading to a significantly increased understanding of \ac{ML} pipelines produced by \ac{AutoML} and the \ac{AutoML} optimization itself.
\end{abstract}



\maketitle

\section{Introduction}
\Acf{ML} has become a vital part in many aspects of daily life. Yet, building well-performing \ac{ML} applications is a challenging and time-consuming task that requires highly specialised data scientists and domain experts. The limited availability of data scientists slows down the further dissemination of \ac{ML}. \Ac{AutoML} aims at improving the current approach of building \ac{ML} applications in two aspects:
\begin{enumerate*}
    \item \ac{ML} experts can benefit from automating tedious tasks, including \ac{HPO}, leading to higher efficiency and greater focus on more challenging tasks; and
    \item domain experts can be enabled to build \ac{ML} pipelines on their own without having to rely on an \ac{ML} expert.
\end{enumerate*}
In the beginning, \ac{AutoML} was only used for a few aspects of the data science endeavour, \eg, tuning the hyperparameters of a classification algorithm. More recently, huge improvements have been made that enable users to automate the complete process from data encoding, preprocessing, and feature engineering to model building. Lately, researchers have proposed many new approaches for \ac{AutoML}, which have reached astonishing performances, \eg, \cite{Feurer2015,Olson2016,Falkner2018,Akiba2019}. With the ever-increasing degree of automation, understanding and validating the behaviour of \ac{AutoML} systems becomes inherently more difficult. Although multiple commercial solutions have been released in the past years, \eg, \cite{Clouder2018,H2O.ai2019,Das2020}, \ac{AutoML} is still primarily an active research topic and only has a niche existence in the larger \ac{ML} universe \cite{GoogleTrends2022}.

Currently, \ac{AutoML} is only seen as a tool to aid human users during their \ac{ML} endeavour in an interactive fashion \cite{Wang2019a}. However, current implementations are not designed with human interaction in mind. From a user's perspective, \ac{AutoML} systems are a black-box that creates another black-box, namely the \ac{ML} pipeline, promising to solve their prediction problem. They behave in a \emph{take-it-or-leave-it} manner without providing users sufficient information about the optimization procedure or the resulting \ac{ML} pipelines. Without sufficient information, users cannot make an informed and accountable decision about whether an \ac{ML} pipeline created via \ac{AutoML} should be used at all. During their optimization procedure, a multitude of \ac{ML} pipeline \emph{candidates} are generated that are all able to solve the given task. However, \ac{AutoML} systems tend to create extremely diverse candidates with no significant performance differences \cite{Zoller2021}. Validating and selecting a model from the abundance of provided solutions is a time-consuming task, even for seasoned \ac{ML} practitioners. For domain experts, also called \emph{subject-matter experts}, with specialized knowledge of a particular area but no expertise in \ac{ML}, this task is practically impossible. Furthermore, modern \ac{AutoML} systems build long and arbitrarily complex \ac{ML} pipelines by selecting from dozens of \ac{ML} primitives---each tunable by multiple hyperparameters---, making it inherently impossible for humans to grasp the complete, high-dimensional search space. In combination with the black-box nature of many \ac{AutoML} systems, it is nearly impossible to understand what is going on during the optimization procedure. This lack of transparency and explainability severely limits the trust in \ac{AutoML} \cite{Li2018,Drozdal2020,Wang2021a,Crisan2021}. Furthermore, this makes the application of \ac{AutoML} for automated decision making in high-risk domains, like healthcare or finance, impossible.

In this article, we propose a new visual analytics tool titled \textit{e\textbf{X}plainable \textbf{Auto}mated \textbf{M}achine \textbf{L}earning (XAutoML)} for analysing and understanding \ac{ML} pipelines produced by \ac{AutoML} systems. This visualization aims to empower all user groups of \ac{AutoML}, namely domain experts, data scientists, and \ac{AutoML} researchers, by
\begin{enumerate*}
    \item making the internal optimization procedure and search space of \ac{AutoML} systems transparent and
    \item providing enough information to validate and select automatically synthesized \ac{ML} models quickly.
\end{enumerate*}
By combining existing techniques from \acf{XAI} and visualizations tailored to \ac{AutoML}, \name{XAutoML} provides a holistic visualization that makes the performed optimizations \textit{transparent} and the synthesized \ac{ML} models \textit{explainable}. Users can compare pipeline candidates, analyse the optimization procedure independent of the actually used \ac{AutoML} system to gain \emph{process} insights, inspect single \ac{ML} models to gain \emph{data} and \emph{model} insights, and inspect \ac{ML} ensembles, which are often produced during the \ac{AutoML} optimization. \name{XAutoML} is integrated with \name{JupyterLab} \cite{JupyterLab2022} to blend with the usual data science workflow. It provides measures to import the result of an \ac{AutoML} optimization and export analytical results directly to \name{Jupyter} for further manual analysis.

The contributions of our work are summarised as follows:
\begin{itemize}
    \item By combining existing visualization techniques with methods from \ac{XAI}, \name{XAutoML} can visualize different aspects of the underlying \ac{AutoML} system as well as the generated \ac{ML} models in a single holistic framework covering data, model, and process insights all at once.
    
    \item In semi-structured interviews, the informational needs of 36 domain experts, data scientists, and \ac{AutoML} researchers for interacting with \ac{AutoML} are gathered. This is the first structured evaluation of requirements of domain experts and \ac{AutoML} researchers for \ac{AutoML} systems at all.
    
    \item We introduce two new visualizations to explain aspects of the \ac{AutoML} optimization procedures for users with minimal \ac{AutoML} knowledge. In combination with improved existing visualizations, \name{XAutoML} is able to visualize all important aspects of the optimization procedure of state-of-the-art \ac{AutoML} systems.
    
    \item We perform a user study with the same diverse user group from the requirements analysis to validate \name{XAutoML}. The study proves that \name{XAutoML} is usable and useful for participants from the three different user groups and helps validating and understanding the \ac{ML} models created by \ac{AutoML}. Participants highlighted the benefits of integrating the visual analytics with \name{JupyterLab}.
\end{itemize}

This article is structured as follows: Section~\ref{sec:related-work} introduces related work. In Section~\ref{sec:requirements-analysis}, requirements for \name{XAutoML} are collected based on usage scenarios, a card-sorting task by potential users and a literature review, followed by the actual design of \name{XAutoML} in Section~\ref{sec:xautoml}. The intended usage of \name{XAutoML} is evaluated in a user study in Section~\ref{sec:evaluations}. This article closes with a discussion in Section~\ref{sec:discussion} followed by a brief conclusion.

\section{Related Work}
\label{sec:related-work}
Our work on visualizing and explaining the decision-making of \ac{AutoML} systems builds upon a significant amount of related work in the areas of \ac{AutoML}, \ac{XAI} and visual analytics for \ac{AutoML}.

\subsection{Automated Machine Learning}
The term \ac{AutoML} summarizes systems and techniques that enable an automated creation of fine-tuned \ac{ML} pipelines with minimal human interaction \cite{Zoller2021}. Those techniques promise to enable domain experts without knowledge of \ac{ML} or statistics to build \ac{ML} pipelines on their own. In addition, data scientists can increase their productivity by automating specific steps of their workflow.

In the beginning, \ac{AutoML} methods covered only single aspects of creating an \ac{ML} pipeline. The earliest works in \ac{AutoML} focused on optimizing the hyperparameters of a single \ac{ML} algorithm \cite{Hutter2011,Bergstra2011}. Specialised systems only consider feature engineering \cite{Lam2017,Katz2017,Chen2018} or the composition of multiple \ac{ML} primitives into complex pipeline structures \cite{Lake2017,Drori2018}. More recently, \ac{AutoML} systems that are capable of synthesising fine-tuned pipelines, including data cleaning, feature engineering, and modeling, have emerged \cite{Feurer2015,Olson2016,Swearingen2017,Mohr2018,Zoller2021a}.

Given an input dataset, a loss function, and a predefined search space, \ac{AutoML} systems use a variety of different strategies to generate \ac{ML} pipelines. For example, \name{auto-sklearn} \cite{Feurer2015} uses Bayesian optimization for algorithm selection and hyperparameter optimization, \name{TPOT} \cite{Olson2016} uses genetic programming for building complex shaped pipelines and \ac{HPO}, \name{ATM} \cite{Swearingen2017} combines multi-armed bandit learning with Bayesian optimization for optimizing the hyperparameters of a fixed pipeline and \name{dswizard} \cite{Zoller2021a} combines \ac{MCTS} for pipeline structure search with Bayesian optimization for \ac{HPO}.

Modern \ac{AutoML} systems incorporate many techniques from standard \ac{ML} to boost the performance of the synthesized end-to-end pipelines, \eg, \cite{Feurer2015,Falkner2018,Alaa2018,Zoller2021}. Those techniques aim to either decrease the optimization duration, like hierarchical search spaces or multi-fidelity approximations, or improve the predictive performance, for example via ensemble learning. Although those techniques are helpful for the optimization, they make the optimization procedure more complex and difficult to understand.

Several recent studies have revealed that data scientists do not trust \ac{AutoML} systems \cite{Wang2019a,Drozdal2020,Wang2021a,Crisan2021}. Even though participants acknowledged that such systems were able to provide high quality solutions \cite{Wang2021a}, they refused to use them as they do not want to be accountable for a model they do not understand \cite{Drozdal2020}. Furthermore, data scientists even argued that \ac{AutoML} should be limited to people with \ac{ML} knowledge, to prevent people from ``automating bad decisions'' \cite{Crisan2021}. Interestingly, participants of all referenced studies named the limited transparency as well as missing explanations of the final \ac{ML} model as the main reasons for their limited trust.

\subsection{Explainable Artificial Intelligence}
\Ac{XAI} is the research area concerned with explaining \ac{AI} systems in a way that can be understood by humans. Usually, automated decision making systems and \ac{ML} models do not operate in a vacuum; rather they, at least to some extent, have to cooperate with human users. For \ac{ML} models to be relevant and helpful, humans have to accept decisions made by those models. Yet, humans have the desire to understand a decision or get an explanation because they do not tend to trust decisions made by others blindly. This directly conflicts with the \emph{black-box} nature of modern \ac{ML} models and \ac{AutoML} systems \cite{Burkart2021}. In the following, several \ac{XAI} techniques are introduced shortly.

\ac{ML} can be restricted to models that are inherently \emph{interpretable}, meaning that the reasoning of a model as a whole can be understood by humans in a reasonable time \cite{Lipton2018}, removing the black-box nature of \ac{ML} completely. This requires the model to be \emph{transparent} and the mapping of data inputs to predictions to be comprehensible \cite{Doran2017}. Unfortunately, interpretable models, like linear models, usually perform worse than black-box models like \acp{SVM} or artificial neural networks, implying a trade-off between explainability and accuracy \cite{Freitas2019,Burkart2021}.

Instead of relying on interpretable models, surrogate approaches explain an arbitrary black-box model by producing similar predictions using an interpretable model. Global surrogates are trained to approximate a black-box model on the complete input space \cite{Molnar2019}, for example approximating an artificial neural network with a decision tree \cite{Schaaf2019}. In contrast, local surrogates are only valid for single data instances and their direct vicinity. Consequently, only local insights of a model can be obtained. \name{LIME} \cite{Ribeiro2016,Ribeiro2018,Ribeiro2020} explains a single prediction of an arbitrary \ac{ML} model by generating artificial data instances in the neighborhood of the selected prediction and fitting a linear, interpretable model to this local dataset.

As an alternative to explaining the black-box model, \ac{XAI} techniques can also be used to explain the relation of input features to the dependent variable. \Acp{PDP} visualize the average relation of a set of input features to the target variable marginalised over all other features \cite{Friedman2001}. Similarly, \acp{ICE} visualize the marginal performance of single-data instances separately \cite{Goldstein2015}. Finally, permutation feature importance \cite{Breiman2001} measures the impact of single features on the predictive power of an \ac{ML} model. By randomly shuffling the values of a single feature, the relation of the feature with the dependent variable is broken. As a consequence, important features induce a large accuracy decrease of the \ac{ML} model while unimportant features should not influence the accuracy at all.

Applying \ac{XAI} techniques to \ac{AutoML} has been tested in a few publications. \citet{Freitas2019} suggests restricting \ac{AutoML} to only interpretable models and shows that this limitation only induces non-significant performance decreases. Yet, when complete end-to-end pipelines are synthesized, having an interpretable model does not explain the complete pipeline. \name{AutoPrognosis} \cite{Alaa2018} creates explanations of the final model using decision lists as a global surrogate. Finally, \name{Amazon SageMaker Autopilot} \cite{Das2020} makes the optimization procedure more transparent by exporting ready-to-use \name{Jupyter} notebooks containing models tested during the optimization.

\subsection{Visual Analytics for AutoML}
In the context of \ac{AutoML}, two different groups of visual analytics tools exist:
\begin{enumerate*}
    \item tools for explaining the \ac{AutoML} optimization procedure and
    \item tools for assisting a user with selecting one of the constructed \ac{ML} models.
\end{enumerate*}
Besides presenting new visual analytics tools for \ac{AutoML}, recent studies have analysed the desired interactions between \ac{AutoML} systems and human users. Those studies are analysed in more detail in the requirements analysis in Section~\ref{sec:requirements-analysis}.

\paragraph{Explaining the AutoML Optimization}
Some \ac{AutoML} systems provide basic visualizations about the optimization procedure to offer some degree of transparency. These include the performance of all tested candidates over time \cite{Feurer2015,Akiba2019} and a parallel coordinate view \cite{Inselberg1990} to visualize all evaluated hyperparameters for single models \cite{Golovin2017,Liaw2018,Akiba2019,Liu2019}. Alternatively, other frameworks provide simple visualizations regarding data insights like feature importance, local explanations, or data distributions \cite{Erickson2020,mljar2021}. Although those methods allow the inference of valuable information for experienced users, many important details, like any information about the behaviour of the constructed pipelines, are missing.

While parallel coordinates are easy to understand for a single algorithm with a few hyperparameters, they can become unreadable for complete pipelines as dozens of different hyperparameters, scattered across multiple pipeline steps, are often present at once \cite{Weidele2019}. \name{AutoAIViz} \cite{Weidele2020} introduces \ac{CPC} to visualize hyperparameters of complete sequential pipelines produced by \name{AutoAI} \cite{Wang2020}. By stacking parallel coordinates hierarchically, users are only presented with a limited number of axes at once, namely one axis for each step in the pipeline. Stacked axes can be expanded to reveal individual hyperparameters. \name{CAVE} \cite{Biedenkapp2019} provides post-hoc visualizations for \name{SMAC} \cite{Hutter2011} to analyse selected aspects of the generated \ac{ML} pipeline. \name{ATMSeer} \cite{Wang2019} provides both transparency and controllability for \name{ATM} \cite{Swearingen2017}, the underlying \ac{AutoML} system. Users can observe the optimization progress---selecting a classifier and optimizing its hyperparameters---and the performance of the generated \ac{ML} models in real-time with different granularity. In addition, users can control the optimization procedure and adjust the search space in-place during an optimization. \name{Hypertendril} \cite{Park2019,Park2021} focuses on visualizing the \ac{HPO} search strategy of \ac{AutoML} systems. By combining a parallel coordinates view with a scatter plot of sampled values of a limited set of hyperparameters over time, users are able to distinguish different search strategies. In addition, \name{Hypertendril} provides an estimate of the hyperparameter importance to guide users while adjusting the search space. All aforementioned visualization tools require a fixed pipeline structure. In contrast, \name{PipelineProfiler} \cite{Ono2021} specializes on comparing different pipeline structures visually. A comprehensible visualization of different pipelines structures is provided by merging multiple structures into a single \ac{DAG}. In addition, \name{PipelineProfiler} provides the option to render the visualization directly in \name{Jupyter}. \name{Optuna} \cite{Akiba2019} provides an interactive visualization to provide insights into the \ac{AutoML} optimization process by visualizing evaluated hyperparameter values, relationships between hyperparameters or feature importance. Similarly, \name{NNI} \cite{nni2021} provides visualizations of constructed neural network architectures. In contrast, \name{AutoWeka} \cite{Kotthoff2016} provides an interactive exploration of the training data with regards to data distributions.

While those visualization tools provide valuable insights, there are still severe limitations: All presented frameworks visualizing the \ac{AutoML} optimization process can either handle different pipeline structures or provide insights for hyperparameter optimization, but considering both aspects simultaneously is crucial for modern \ac{AutoML} systems. Other frameworks provide explanations with focus on the input data. Yet, no visual analytics tool covers both aspects, data insights and \ac{AutoML} process explanations, at the same time. Other important aspects like \ac{ML} ensembles are completely ignored. Furthermore, visualizations are often only post-hoc static figures and do not provide an option for users to interactively retrieve desired information. While the integration of a visual analytics tool with a single \ac{AutoML} system enables detailed inspections and control over this system, it basically prevents the application to other \ac{AutoML} systems. An \ac{AutoML} visualization should be compatible with multiple \ac{AutoML} systems to be relevant for a wide user base. With \name{XAutoML} we aim to overcome these limitations.

\paragraph{Visual Analytics for Model Selection}
As \ac{AutoML} systems produce numerous different models with very similar performances during their search procedure, assessing the performance and finally selecting a model is a challenging task for users. Various visual analytics systems have been proposed to assist domain experts with model selection: \name{EMA} \cite{Cashman2019}, \name{ClaVis} \cite{Heyen2020}, \name{Visus} \cite{Santos2019}, and \name{Boxer} \cite{Gleicher2020} provide a ranking of evaluated candidates and options to compare the performance of multiple models using different metrics and confusion matrices. While \name{Boxer} especially focuses on the analysis of \ac{FATE}, \name{EMA} and \name{Visus} provide a front-end to guide domain experts during initial exploratory data analysis and specifying an \ac{AutoML} optimization procedure. \name{ClaVis} focuses on comparing models using different scores like the performance or used hyperparameters. Even though those visual analytics tools provide simple performance statistics of \ac{ML} models, no explanation of the actual model behaviour in terms of \ac{XAI} is provided to aid users in selecting \ac{ML} models for high-risk domains requiring further model reasoning \cite{Gil2019a}. \name{explAIner} \cite{Spinner2019} combines explanation techniques in a single interactive user interface allowing users to explain various aspects of an \ac{ML} model. Yet, it is limited to evaluating a single model at once making it unsuited for \ac{AutoML}.

Our approach not solely aims at assisting users in selecting a well-performing model, but it is also intended to provide enough information to validate the selected model. This implies that model explanations should be considered as an additional factor for model selection besides the pure model performance. As those visual analytics tools are exclusively targeted on domain experts, crucial information about the \ac{ML} model, like the pipeline structure, are hidden on purpose, making those tools less suited for experienced \ac{ML} users. With \name{XAutoML}, we aim to overcome these limitations allowing users to select the aspects of a model analysis they are interested in. Finally, all mentioned papers failed to actually collect potential requirements from domain experts. We aim to fill this gap by performing a requirements analysis prior to designing \name{XAutoML}.

\paragraph{Commercial AutoML Systems}
Besides the previously discussed open-source \ac{AutoML} systems, many commercial tools for end-to-end \ac{AutoML} have been created in recent years, \eg, \name{Amazon SageMaker Studio}\cite{Das2020}, \name{Azure ML Studio} \cite{AzureML2022}, \name{Dataiku} \cite{Dataiku2023}, \name{DataRobot} \cite{DataRobot2021}, \name{Google Cloud AutoML} \cite{Golovin2017}, or \name{H2O.ai}\cite{H2O.ai2019}. These tool usually offer an interactive user interface covering the complete process from data ingestion to model deployment. Consequently, visualizations aiming to validate models and to understand the \ac{AutoML} optimization process are also available.

Basically all commercial tools provide visualizations for gaining data insights like data distributions or feature importance. In addition, information about the performance of all created models is presented with varying degrees of details. Yet, technical information like used hyperparameters, the underlying search spaces, or information how the \ac{AutoML} optimization process itself is actually executed is often not given. A potential explanation could be that commercial \ac{AutoML} tools often target users with no \ac{ML} expertise. Yet, this prevents users with prior knowledge in \ac{ML} or \ac{AutoML} from gaining valuable insights. Furthermore, due to the closed-source nature of these commercial tools users are limited to the provided visualizations. It is not possible to dig deeper into specific areas as these tools provide a rather restrictive interface for data extraction.

While we are not able to create visual analytics with comparable ease of use and seamless integration as these commercial tools offer, we still aim to provide additional value. Namely, we want to improve two main issues:
\begin{enumerate*}
    \item Missing information about the underlying \ac{AutoML} process hinders users with existing \ac{ML} and \ac{AutoML} expertise. Visual analytics should cover all important aspects of \ac{AutoML}.
    \item The visual analytics system should be open to allow users to create their own visualizations on demand.
\end{enumerate*}

\section{Requirements Analysis}
\label{sec:requirements-analysis}

We want to understand the needs for visual analytics of \ac{AutoML} practitioners. Therefore, we first envision three prototypical usage scenarios for visual analytics in the context of \ac{AutoML} based on a literature review. To validate these scenarios, we collect the requirements from different user groups through a card-sorting exercise. Based on these results, we identify commonly requested explanations when using \ac{AutoML} systems and distill them into three key analytical needs that visualization can solve. Finally, we formulate four design goals for \name{XAutoML}.

\subsection{Usage Scenarios}
\label{sec:case-studies}

Various studies have collected potential requirements from data scientists regarding the use of \ac{AutoML} \cite{Drozdal2020,Liao2020,Wang2021a,Wang2021}. The important message from all these studies is that participants refused to use an \ac{ML} model constructed by \ac{AutoML} just because it performed well. Instead, further insights to validate the \ac{ML} model were requested, \eg, to explain the behaviour of a model due to legal constraints. In addition, more information about the optimization procedure was desired, including information about other evaluated \ac{ML} models and the internal reasoning of the optimizer. To highlight how \name{XAutoML} could support users of \ac{AutoML} systems, we envision how it can be used in combination with a prototypical \ac{AutoML} system under ideal settings for three real-world usage scenarios covering the aforementioned requested information. Potential limitations induced by the used \ac{AutoML} system are not considered.

\subsubsection{Scenario 1: Understanding the Generated ML Pipelines}
In this scenario, we illustrate how a domain expert persona named Alice, a physician with no \ac{ML} expertise, can use \name{XAutoML} to analyse models generated via \ac{AutoML}. Alice aims to create a model to predict whether a patient has a high risk for a cardiovascular disease. With no prior knowledge of \ac{ML}, she uses a web-based service to fit different classification models on historical records of patients she has collected. After the optimization is done, Alice is presented with a final model and the \ac{AutoML} systems reports a validation accuracy of \(87\%\).

Due to the high-risk nature of misclassifications, Alice does not want to use the suggested best-performing model blindly but instead wishes to examine the produced models closely before using them. She loads the results in \name{XAutoML} to gain more insights. At a first glance, she notices that all displayed models in the leaderboard have a similar performance. She opens the performance details for the best performing model. The first important information is the specificity and selectivity of the model. Alice reviews the confusion matrix to assess the number of misclassifications. After checking that the other candidates yield similar specificity, she continues to validate that the model makes sensible decisions. Therefore, she opens the feature importance inspection and global surrogate. Alice notices that most models use an \ac{EMG} diagram shape as the most important feature, which aligns with her expectation. The best candidate uses the patient's sex as the second most important feature to classify a high risk. She disagrees with this assessment and moves on to the second-best candidate. This model uses the presence of exercise-induced angina, which makes more sense in her opinion, so she decides to further analyse this candidate. Next, Alice takes a closer look at a few patient records. By selecting records with low confidence, Alice analyses why the model is not sure about the predicted class. By comparing the provided local surrogates with her background knowledge, she verifies that these particular patients have indeed no clear symptoms. Satisfied with the explanations, Alice decides to pick the second-best model\footnote{
    For the sake of completeness, most \ac{AutoML} systems still lack the option to actually deploy models to production without programming skills. Yet, this is a technical problem that is solved by many commercial solutions, \eg, \name{H2O.ai} \cite{H2O.ai2019} or \name{DataRobot} \cite{DataRobot2021}, and is not further considered in this case study.
}.

\subsubsection{Scenario 2: Steering the Optimization Procedure}
Next, we illustrate how \name{XAutoML} helps a data scientist persona to evaluate the search process as a whole and inspect single pipeline candidates. Bob, who is a consultant developing \ac{ML} models for customers, is tasked with creating an optimized \ac{ML} pipeline. As usual, he starts up \name{JupyterLab} to perform an exploratory data analysis. After familiarising himself with the dataset, Bob wants to create a set of baseline models using \ac{AutoML}. He starts the optimization procedure in the \ac{AutoML} system of his choice. After the optimization procedure has finished, he loads the results in \name{XAutoML}.

Bob first scans the leaderboard with all candidates and observes that the accuracy differences between the top pipelines are less than 2\%. By comparing the different pipeline structures, Bob can quickly examine which algorithms are used in the single models. Before exploring a model in more detail, he wants to examine whether the search space was sufficiently investigated. He switches to the search space overview and examines the explored pipelines. At first, he notices that only a single pipeline using an \ac{SVM} was tested. A glance at the optimization progress view shows that all candidates are quite evenly scattered in the search space and no cluster of similar candidates was created. With this information, Bob decides that the search process was prematurely stopped and should be resumed for some more time. While checking the search space overview, he also notices that pipelines using a decision tree performed worse than the remaining candidates. Therefore, Bob decides to remove decision trees altogether. After the second optimization run, he checks the optimization progress view again and this time a cluster of similar candidates is visible. Convinced that the search procedure was performed thoroughly, he again turns to inspecting single candidates.

By selecting a well-performing pipeline from the leaderboard, the details of the according pipeline are revealed. A glance at the hyperparameter importance view reveals that the selected imputation strategy has the largest impact on the pipeline performance. Bob selects the according imputation step in the pipeline visualization graph. He continues to visualize the intermediate data produced by the imputation algorithm and tests the impact of different imputation strategies on the data. Satisfied with the explored hyperparameters, he decides not to start a new optimization procedure.

\subsubsection{Scenario 3: Validating the Behaviour of AutoML Methods}
In the final scenario, a persona named Charlie, who is an \ac{AutoML} researcher, uses \name{XAutoML} to visualize and verify an \ac{AutoML} algorithm under active development. This new algorithm is supposed to build pipeline structures with increasing complexity and length using \ac{MCTS}. For each created pipeline, the hyperparameters are supposed to be optimized via \name{hyperopt} \cite{Bergstra2011}, a well-established model-based \ac{HPO} algorithm. The performance of his approach is fairly good but worse than existing \ac{AutoML} systems. To analyse this shortcoming, Charlie loads a recent optimization run into \name{XAutoML} and opens the search space inspection.

He examines all evaluated pipeline structures. By rewinding through the complete optimization run using a time-lapse function he can examine the traversal of the search space step by step. At first glance, the pipeline construction performs as expected and better pipelines are found over time. Next, he turns to inspecting the evaluated hyperparameters. He selects a pipeline with mediocre performance at random to check the hyperparameters of the used classifier. He checks the sampled values of a single hyperparameter during the complete optimization. Charlie notices that the hyperparameters appear to be sampled at random without converging to a local minimum. \name{hyperopt} uses Bayesian optimization to select promising hyperparameter values. This requires building an internal model of the search space based on previous observations. In the beginning, random values are sampled to build an initial model. The new \ac{AutoML} system simply did not evaluate enough hyperparameters for each pipeline candidate to build an initial model and basically only used random search for \ac{HPO}. Realising this, Charlie decides to increase the number of hyperparameter evaluations of each pipeline candidate to utilize the full potential of \name{hyperopt}.

\subsection{Collecting Informational Needs}
\label{sec:requirements-study}
All studies mentioned in Section~\ref{sec:case-studies} only considered data scientists as potential users of visual analytics systems. To the best of our knowledge, no study has evaluated the requirements of domain experts or \ac{AutoML} researchers for using \ac{AutoML} systems with or without visual analytics. Therefore, we performed a requirements analysis, before designing and implementing our visual analytics tool, to collect important informational needs for \ac{AutoML} from a diverse user base covering domain experts, data scientists, and \ac{AutoML} researchers.

\begin{figure}
    \centering
    
    \begin{subfigure}[b]{0.49\textwidth}
         \centering
         \includegraphics[width=\textwidth]{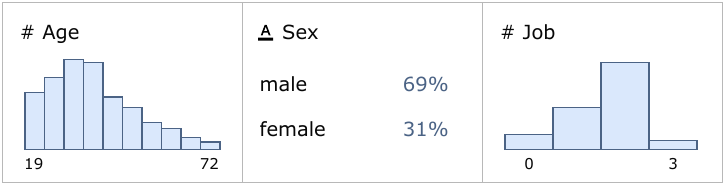}
         \caption{"View statistics of input data".}
         \Description{Visualization of data statistics. Displayed are two numerical features as histograms and the distribution percentages for a categorical feature.}
     \end{subfigure}
     \hfill
     \begin{subfigure}[b]{0.49\textwidth}
        \centering
        \includegraphics[width=\textwidth]{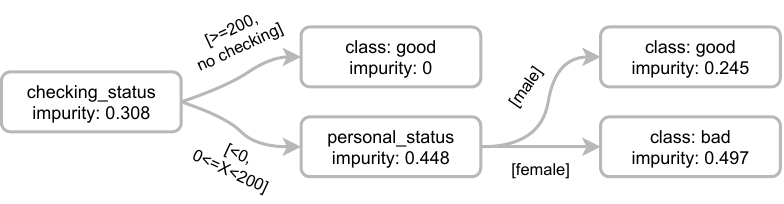}
        \caption{"View global surrogate for model".}
        \Description{Visualization of a global surrogate. Displayed is a decision tree with five nodes to approximate an arbitrary black-box model. Each node contains either the feature used for this data split or in case of a leave node the majority class. Edges contain the conditions used for splitting the data.}
     \end{subfigure}
    
    \caption{Exemplary visualizations of two cards in the card-sorting task.}
    \label{fig:card_examples}
\end{figure}

In our study, we conducted a card-sorting exercise to understand informational needs in visual analytics for \ac{AutoML}. Participants were given a set of digital cards containing explanations of and information about various parts of the \ac{AutoML} optimization procedure. Each card contained a verb, \ie~"view", "know", and "compare", followed by a single piece of information related to either the \ac{AutoML} system itself or the produced \ac{ML} models. For example, one card stated "View statistics of input data" and another stated "View global surrogate for model". As the cards may contain concepts unknown to the participants, each card was accompanied by an example visualization and textual description. Figure~\ref{fig:card_examples} contains the according visualizations for the previous examples. In addition, participants were encouraged to ask questions about unclear cards.

We created \(24\) different cards based on the usage scenarios, similar studies for data scientists \cite{Drozdal2020,Liao2020} and requested information in interviews with \ac{AutoML} users \cite{Crisan2021,Wang2021a}. These cards, denoted as R01 to R24, covered various aspects of \emph{data} (examining the processed data), \emph{process} (understanding the \ac{AutoML} procedure), and \emph{model} (explaining the \ac{ML} model) insights. The complete set of cards and the raw results are available in Table~\ref{tab:card_sorting_results} in the Appendix. Besides the predefined digital cards, participants were encouraged to add cards with additional information or explanations that they would be interested in. Participants were asked to rank the available cards in a single list from most to least \emph{important for establishing trust} in \ac{AutoML} and the models produced by it. In addition, participants had the opportunity to discard cards as irrelevant. These cards are inserted at the end of the list with the average rank of all irrelevant cards.

For this study 36 participants with diverse backgrounds were recruited. Participants were acquired via a snowball sampling method, beginning with colleagues involved in data science, fellow researchers, contacts from research projects and company contacts. A total of \(44\%\) of the participants assigned themselves the role data scientist, \(31\%\) domain expert, and \(25\%\) \ac{AutoML} researcher. However, many participants also stated that a clear assignment to just one of these roles is difficult. A total of 44\% of the participants worked in academia with 56\% working in industry. The participants' backgrounds can be further broken down into different professions: academia (25\%), information technologies (22\%), healthcare (17\%), manufacturing (14\%), robotics (8\%), finance (8\%), automotive (3\%), and business administration (3\%). In addition, participants were asked to rate their prior knowledge in \ac{ML} and \ac{AutoML} on a scale from 1 (very little) to 5 (very much). On average participants have an \ac{ML} expertise of \(3.00 \pm 1.39\) with several participants having no expertise with \ac{ML} at all. Similarly, results for \ac{AutoML} expertise (\(2.56 \pm 1.46\)) were also quite diverse with many participants having never heard of \ac{AutoML} before. More information is available in Table~\ref{tab:prior_knowledge} in the Appendix. In general, the group of participants was quite diverse with some participants having no prior experience using \ac{ML} or \ac{AutoML} and other participants being involved in either \ac{ML} or \ac{AutoML} or even both for many years.

Prior to the study, participants were informed about the nature of the study and its procedure. For the study itself, participants were invited to a roughly 20 minutes long online interview with one of the authors. First, demographics of the participants were collected. Next, as we did not expect all participants to have prior experience with \ac{AutoML}, we provided a baseline \ac{AutoML} interface based on screenshots of a commercial \ac{AutoML} system excluding easily identifiable features like the company's logo or name. The screenshots were presented and explained by one of the authors and covered the complete \ac{AutoML} process from data import, manually selecting relevant features and the target feature, starting the optimization process, waiting for it to finish, and finally inspecting basic performance metrics of the created model. We also provided an introduction to \ac{AutoML} from an end-user perspective while presenting the various screenshots. The goal was to familiarise participants with the high-level design goals of \ac{AutoML} and to generate a general understanding of how \ac{AutoML} can aid them in their work. The session was concluded by the actual card-sorting task using a collaborative website for brainstorming with sticky notes.

\begin{figure}
    \centering
    \includegraphics[width=\textwidth]{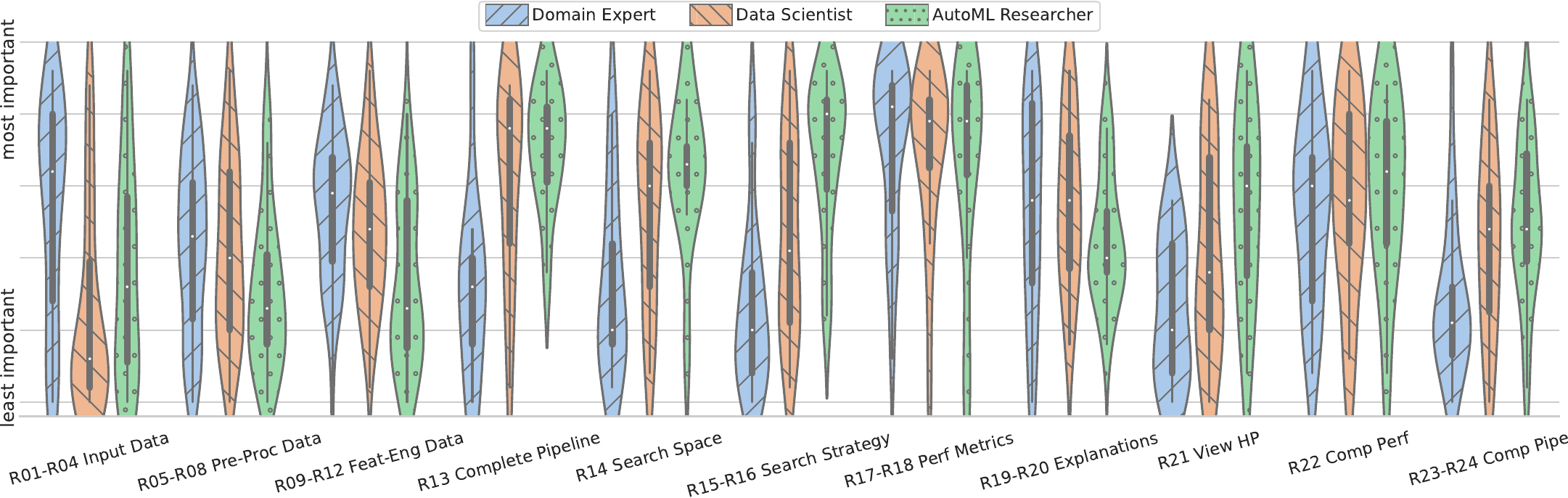}
    \caption{Distribution of the ranking of the different information from most to least important for \emph{establishing trust in \ac{AutoML}}. For a better visualization, some related cards are combined. The complete set of cards, including the raw results, is available in Table~\ref{tab:card_sorting_results} in the Appendix.}
    \label{fig:results_card_sorting}
    \Description{Results of the card sorting task. Fully described in the text and in Table~\ref{tab:card_sorting_results}.}
\end{figure}

Although participants were given the opportunity to add their own cards, only seven decided to use this option. Therefore, we first discuss the results of the pre-defined cards before considering the additional cards in more detail. Figure~\ref{fig:results_card_sorting} provides the ranking of the importance of the \(24\) predefined cards grouped by domain experts, \name{AutoML} researchers and data scientists. As expected, nearly all participants rank information about the final performance of a model, in textual (R17) or visual form (R18), as by far the most important information. Providing explanations for a model's behaviour in form of local (R19) or global surrogates (R20), including the option to compare these values between different models (R22), is the second most important kind of information. Some information, like the ability to view the complete \ac{ML} pipeline (R13) or compare different pipelines (R23), are more important to data scientists and \ac{AutoML} researchers than domain experts. For the majority of cards, no clear preferences were found regarding the importance of the information we presented.

The seven additional cards cover three different topics. Five participants stated that they would like to modify candidates in-place to assess the impact of hyperparameters and different pipeline structures on the performance. Participant P5 expressed their desire to compare the created model with an analytical model, given that such a model already exists. P10 wanted to compare data produced by the different stages in an \ac{ML} pipeline to validate that relevant information is still preserved.

\subsection{Accommodating a Heterogeneous Target Audience}
According to its self-proclamation, \ac{AutoML} targets domain experts and data scientists \cite{Zoller2021}. A third (unintentional) user group is \ac{AutoML} researchers and developers who spend considerable time studying various \ac{AutoML} systems. Figure~\ref{fig:results_card_sorting} shows that a clear distinction between the informational needs of domain experts, \name{AutoML} researchers and data scientists is not possible for most information. Even though slight differences between the three user groups are visible, those differences are often negligible or overlap significantly. Table~\ref{tbl:desired_information} assigns the potential information to the user groups, given that significant differences in the importance exist. Only for eight cards, a significant difference (\(p \leq 0.05\)) can be observed: Information about input data (R02 and R03) is more relevant for domain experts. Data scientists stated they would always perform an exploratory data analysis before building models with \ac{AutoML}. Therefore, they do not require this information in an \ac{AutoML} visualization. Furthermore, domain experts are significantly less interested in understanding and comparing pipeline structures (R13 and R23) and checking hyperparameters (R21) in comparison to the other two user groups. They primarily stated that knowing which algorithms are used in a pipeline would not be helpful as the background knowledge to interpret this information is missing. Finally, \name{AutoML} researchers were significantly more interested in a search space overview (R14) and search strategy visualizations (R15 and R16). For the remaining cards, the informational need does not correlate with one of the three roles but highly depends on the knowledge background of the person. Similar results were also observed in other studies \cite{Drozdal2020,Crisan2021,Wang2021a}.

\begin{table}[t]
    \centering
    \caption{Potential informational needs with significant importance differences between domain experts (DE), data scientists (DS) and \ac{AutoML} researchers (AR). Each user group in the first row is compared with the remaining two groups in the second row. It is listed whether the information is significantly more (+) or less (-) important for the user group in the first row.}
    \label{tbl:desired_information}
    
    \begin{tabularx}{\textwidth}{lX ll ll ll}
        \toprule
        \multicolumn{2}{l}{Information}   & \multicolumn{2}{l}{DE}  & \multicolumn{2}{l}{DS}  & \multicolumn{2}{l}{AR} \\
            &                             & DS  & AR \hspace*{2em}  & DE & AR \hspace*{2em}   & DE & DS\\
        \midrule
        R02 & View the meanings of columns in the raw input data    & +   & +     & -  &      & -  &   \\
        R03 & View statistics of raw input data                     &     & +     &    &      & -  &   \\
        R13 & View the complete processing pipeline                 &     & -     &    & -    & +  & + \\
        R14 & Know what the search space looks like                 &     & -     &    &      & +  &   \\
        R15 & Know how pipelines are chosen                         &     & -     &    & -    & +  & + \\
        R16 & Know how hyperparameters are chosen                   &     & -     &    & -    & +  & + \\
        R21 & View hyperparameters of model                         &     & -     &    &      & +  &   \\
        R23 & Compare differences between pipelines                 &     & -     &    & -    & +  & + \\
        \bottomrule
    \end{tabularx}

\end{table}

Current \ac{AutoML} visual analytics tools always distinguish between either a domain expert, with no expertise in \ac{ML} or programming, and a data scientist, who is only interested in technical model details but not the underlying domain \cite{Cashman2019,Santos2019,Gil2019a,Heyen2020,Gleicher2020,Wang2019,Weidele2020,Park2021,Ono2021}. We argue that the user basis is more diverse, with many shades between the classic domain expert and data scientist. Consequently, the visual analytics tool should not force users into either of the roles; rather, it should enable them to fully use their potential: A domain expert with no \ac{ML} skills but knowledge in data visualization should be able to build their desired visualizations instead of being limited to the visual interface optimized for users with no technical experience. Instead of this one dimensional distinction of user groups, we propose to introduce three orthogonal \emph{knowledge dimensions} related to the usage of \ac{AutoML}:
\begin{enumerate*}
    \item Domain expertise for validating the predictions of a model,
    \item \ac{ML} expertise for model improvement and behaviour validation, and
    \item \ac{AutoML} expertise for optimization refinement and debugging.
\end{enumerate*}
As users can be proficient in multiple knowledge dimensions, eight different user groups can be deduced. It is important to note that all dimensions are continuous and no clear distinction between the eight user groups exists. Instead of creating dedicated visualizations for eight fluent, not clearly separable user groups, a visual analytics tool for \ac{AutoML} should provide, to varying extents, useful information for all knowledge dimensions and users should be able to select relevant information.

For the context of this work, 53\% of all participants have domain expertise, 75\% \ac{ML} expertise, and 28\% \ac{AutoML} expertise with 45\% having only one proficiency and 55\% having at least two proficiencies. For the remainder of this article we will refer to all participants being proficient in domain expertise as domain experts, independent of their other knowledge dimensions. The same holds true for data scientists and \ac{AutoML} researchers.

\subsection{Blending with the Data Science Workflow}
\label{sec:human-in-the-loop}

\begin{figure}
    \centering
    \includegraphics[width=0.9\textwidth]{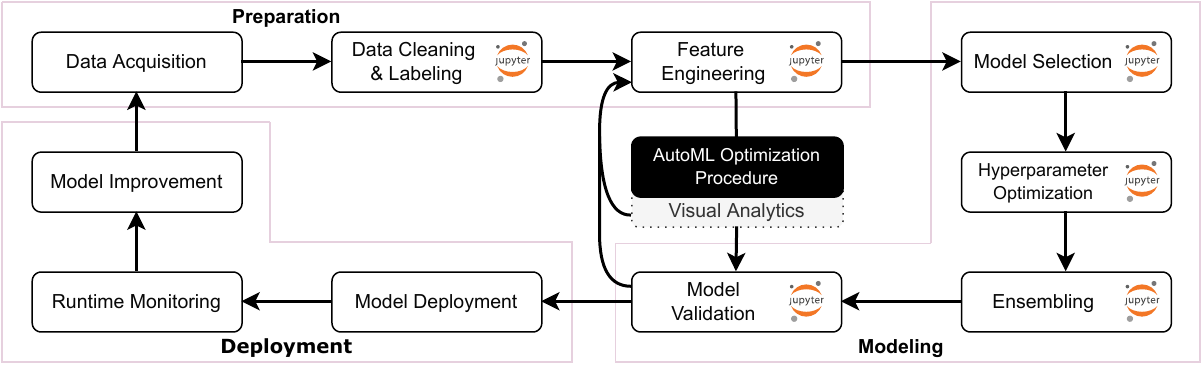}
    \caption{Visualization of the data science workflow, adapted from \cite{Wang2019a}. Steps usually performed in \name{Jupyter} are marked by the \name{Jupyter} logo. \ac{AutoML} provides an alternative path for most of the modeling steps. The intended place of the visual analytics tool is shown in gray.}
    \label{fig:automl_workflow}
    \Description{The usual data science workflow including a potential place for AutoML and the visual analytics. The ten different steps in the data science workflow are displace as circle and AutoML provides a shortcut between the steps feature engineering and model validation allowing users to skip three steps in the modeling phase. The visual analytics is intended as a direct extension to AutoML. In addition to the original data science workflow an additional branching from the visual analytics and model validation to feature engineering is added to indicate that modeling is an iterative procedure.}
\end{figure}

To ease the use of the visual analytics tool, \name{XAutoML} should blend with the typical data science workflow. \citet{Wang2019a} proposed the data science workflow shown in Figure~\ref{fig:automl_workflow}. This workflow contains three stages: preparation, modeling, and deployment, which can be further divided into 10 steps from data acquisition to runtime monitoring and model improvement. The execution of the steps data cleaning to model validation is usually an open-ended, exploratory and iterative process \cite{Batch2018,Cashman2019,Muller2019} which uses ad-hoc visual analytics extensively \cite{Wang2019a}. These visualizations enable practitioners to gain insights into their data and models quickly. In recent years, \name{Jupyter} has become the de-facto standard environment for such interactive visualization tasks \cite{Rule2018,Drozdal2020}.

\name{Jupyter Notebook} \cite{Kluyver2016} and its successor, \name{JupyterLab} \cite{JupyterLab2022}, are web-based programming environments that support interactive data science and scientific computing. \name{JupyterLab} is built around the idea of computational notebooks, which combine code, execution results, and descriptive texts into a single file \cite{Rule2018}.

\ac{AutoML} systems aim to (partially) automate the steps from feature engineering up to ensembling in the data science workflow. If \ac{AutoML} would be more commonly used, it is reasonable to assume that it would be mainly used in the context of \name{Jupyter} as all surrounding steps in the workflow are also mostly executed in it. Instead of providing the visual analytics as a stand-alone external tool, it should be integrated with \name{Jupyter} to provide a cohesive environment for users \cite{Crisan2021}. This integration has an additional advantage: Data analytics range from simple statistical analysis to advanced \ac{ML} techniques. It is virtually impossible for any single visual analytics tool to cover the complete range of possible analytics. Instead of pursuing this unreachable goal, we argue to provide easy options for experienced users to extend the visual analytics with their own ad-hoc visualizations in \name{Jupyter}. To encourage such behaviour, the tool should be designed as a \emph{data pipe} in the data science workflow. In contrast to a \emph{data sink} that only consumes data, namely the results of an \ac{AutoML} optimization, a data pipe is able to emit new data the user can continue working with. Therefore, options for exporting (intermediate) datasets and \ac{ML} artifacts should be implemented. Consequently, \name{XAutoML} could blend seamlessly with the data science workflow, as displayed in Figure~\ref{fig:automl_workflow}.

Although \name{Jupyter} is a mighty tool for data scientists, programming skills are required to create notebooks, which makes it to a certain degree unsuitable for users with no programming skills. Following \citet{Grappiolo2019}, we argue that \name{Jupyter} can become a useful tool for these users if additional visualization features are incorporated into notebooks that enable them to use \name{Jupyter}. Fortunately, \name{JupyterLab} provides a powerful extension \ac{API} that allows the inclusion of interactive JavaScript applications directly into notebooks. These extensions provide the opportunity to eliminate the prerequisite of programming skills, making \name{Jupyter} a usable environment for all types of users.

\subsection{Visualization Needs and Design Goals}
Based on the results from the initial requirements analysis and the literature review, we formulate three visualization needs, denoted as N1 to N3, that \name{XAutoML} aims to support. In addition, four non-functional design goals, denoted as G1 to G4, are listed.

\paragraph{N1. Effective and Efficient Validation of ML Models}
The ability to understand and validate a model produced by \ac{AutoML} is crucial for the prevalence of \ac{AutoML}. If users decide not to use it due to missing trust in the results \cite{Wang2019a}, \ac{AutoML} will fail to become relevant. Therefore, the primary goal of \name{XAutoML} is to provide necessary information to quickly assess the validity of single models and support users in selecting a model from the multitude of produced ones.

\paragraph{N2. Understanding and Diagnosing of AutoML Methods}
\ac{AutoML} methods are currently often designed as black-box optimizations. To provide a better understanding and interpretation of the complex and diverse optimization strategies in \ac{AutoML}, we aim to provide an effective and intuitive visualization of
\begin{enumerate*}
    \item what the complete search space looks like,
    \item how pipeline structures are synthesized, and
    \item how hyperparameters are optimized.
\end{enumerate*}
This visualization shall provide a rough understanding of the underlying search algorithm, without requiring extensive knowledge of \ac{AutoML}.

\paragraph{N3. Search Space Refinements}
As shown in Section~\ref{sec:human-in-the-loop}, \ac{AutoML} is usually used in an iterative workflow. Between two consecutive runs of \ac{AutoML}, users have the option to adapt the underlying search space that is provided as a preset before performing the optimization. To effectively choose a refined search space, users should have information about which regions of the search space perform well and which regions can be pruned.

\paragraph{G1. Align with the Target Audience of AutoML}
\ac{AutoML} is aimed to assist domain experts and data scientists. Therefore, the visual analytics should also target these two groups. In addition, \ac{AutoML} researchers should be able to extract useful information from the visual analytics.

\paragraph{G2. Blend with the Usual Data Science Workflow}
We expect visual analytics to be constantly involved in the workflow of using \ac{AutoML}. Therefore, it is crucial that the visual analytics blend seamlessly with the usual data science workflow. On one hand, this implies that output generated by an \ac{AutoML} system has to be transferred easily to the visual analytics tool or that perhaps the visual analytics tool can be used to start an \ac{AutoML} optimization. On the other hand, this implies that it has to be possible to transfer \ac{ML} models back from the visual analytics tool to the usual data science environment, namely \name{Jupyter}.

\paragraph{G3. Always Provide more Detailed Information}
The informational need of users is diverse. Following the idea of \emph{Five Whys} \cite{Serrat2017}, \name{XAutoML} should provide information in a hierarchical fashion to prevent overloading users with unnecessary and undesired information. At the highest level, only very basic information should be available and users should have the option to dig deeper into certain analytics aspects to get more information. Ultimately, the visual analytics tool can not provide all the information required by a proficient user. Therefore, we aim to add options to \emph{break-out} of \name{XAutoML}. This extends G2 by not only exporting selected models but also exporting artifacts related to models, \eg, intermediate datasets or sub-pipelines, for further manual inspection in \name{Jupyter}.

\paragraph{G4. Be Open to the AutoML World}
The \ac{AutoML} landscape is still rapidly developing and constantly shifting. To stay relevant, the visual analytics tool should not be coupled to a specific \ac{AutoML} implementation. This implies that \name{XAutoML} should only depend on generic information provided by \ac{AutoML} systems. Therefore, the common basis for the visual analytics and \ac{AutoML} implementations should be \name{scikit-learn} \cite{Pedregosa2011}, \name{NumPy} \cite{VanDerWalt2011} and \name{pandas} \cite{McKinney2010}, three commonly used libraries for \ac{ML} in \name{Python}. Consequently, the visual analytics tool will be limited to \name{scikit-learn} pipelines for supervised classification on tabular data for now.

\section{XAutoML: Exploring AutoML Optimizations}
\label{sec:xautoml}

\begin{figure}
    \centering
    \includegraphics[width=\textwidth]{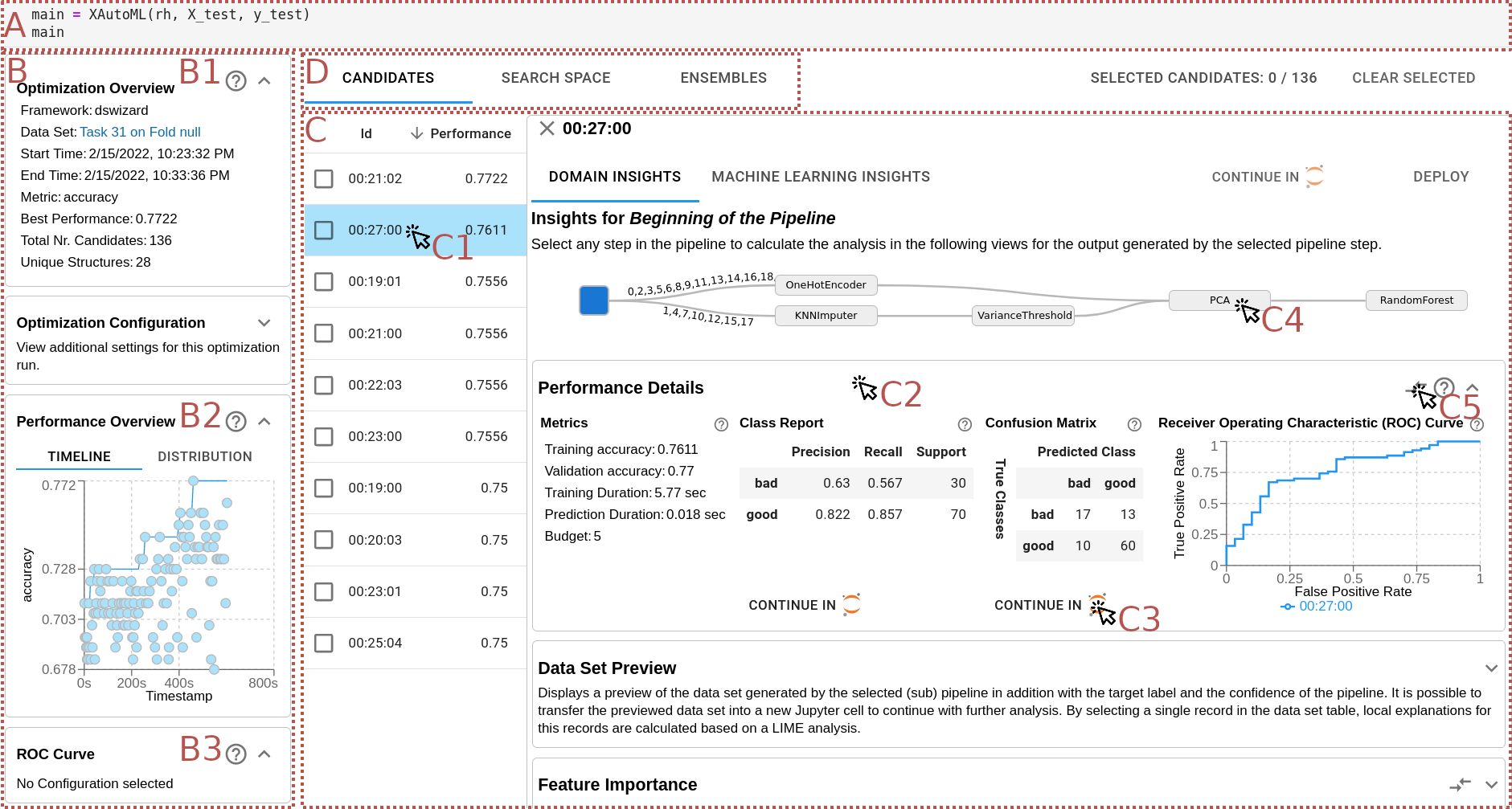}
    \caption{Overview of \name{XAutoML}. The visualization is integrated with \name{Jupyter} and can be accessed with a few lines of code (A). On the left side (B), the optimization overview provides basic statistics about the optimization run (B1), a scatter plot of the accuracy of all candidates over time (B2), and a \ac{ROC} curve of selected candidates (B3, hidden). The \emph{leaderboard} view (C, partially hidden) provides a comprehensive overview of all evaluated candidates. Users can open single candidates (C1) in an overlay on the right-hand side to reveal detailed information about them (only partially visible). The candidate details contain various boxes grouping related information together. In the \emph{performance details} view (C2), performance metrics and basic performance visualizations are available. By clicking the \emph{Continue in \name{Jupyter}} button (C3), the according information can be exported to a new \name{Jupyter} cell. Users can access the \emph{search space} and \emph{ensemble inspection} via the tabs at the top (D).}
    \label{fig:startpage}
    \Description{Screenshot of optimization overview and candidate inspection rendered in Jupyter. Fully described in the figure caption and in the text.}
\end{figure}

To fulfill the requirements, visual needs, and design goals identified in the previous section, we developed \name{XAutoML}. The interface of \name{XAutoML} is embedded in \name{Jupyter} and consists of an \emph{optimization overview}, a \emph{candidate inspection} to explore individual pipeline candidates, a \emph{search space inspection} to gain insights about the optimization process and search space, and finally, an \emph{ensemble inspection} view. The individual views are described in more detail in the following sections. Figure~\ref{fig:startpage} provides an overview of \name{XAutoML}.

\subsection{Optimization Overview}
The \emph{optimization overview} (Figure~\ref{fig:startpage}, B) provides an overview of the results of an \ac{AutoML} optimization at a high level, including the total optimization duration, the number of evaluated candidates, and the performance of the best candidate. In addition, basic performance visualizations, namely the performance of all candidates over time and the number of candidates grouped by performance, are available. Users can select multiple candidates in the overview to compare their \ac{ROC} curves. With these views, users can quickly identify interesting models for a detailed inspection. The candidate inspection view of interesting models can be opened directly from the overview.

\subsection{Validating Machine Learning Models}
The \emph{leaderboard} (Figure~\ref{fig:startpage}, C) provides a tabular overview of all evaluated candidates. Each row displays the candidate id, performance (R17), average prediction time (currently not visible), and the used classifier (currently not visible) of a single candidate. Via the \emph{Continue in \name{Jupyter}} button (currently not visible), the corresponding candidate can be exported to \name{Jupyter}. The leaderboard is intended to provide a rough first impression of the optimization with the option to select single candidates for further inspection.

\begin{figure}
    \centering
     \begin{subfigure}[b]{0.49\textwidth}
        \includegraphics[width=\textwidth]{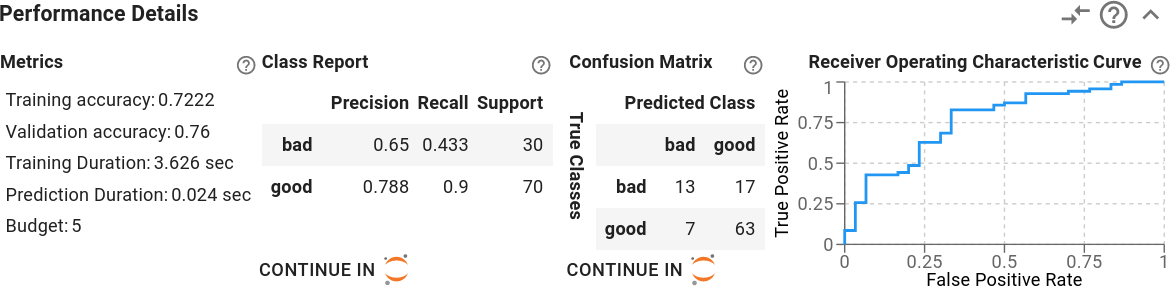}
        \caption{Performance details. From left to right some basic metrics, a class report, a confusion matrix and a \ac{ROC} curve is displayed.}
        \label{fig:performance_details}
        \Description{Screenshot of performance details view. Fully described in the text.}
    \end{subfigure}
    \hfill
    \begin{subfigure}[b]{0.49\textwidth}
        \includegraphics[width=\textwidth]{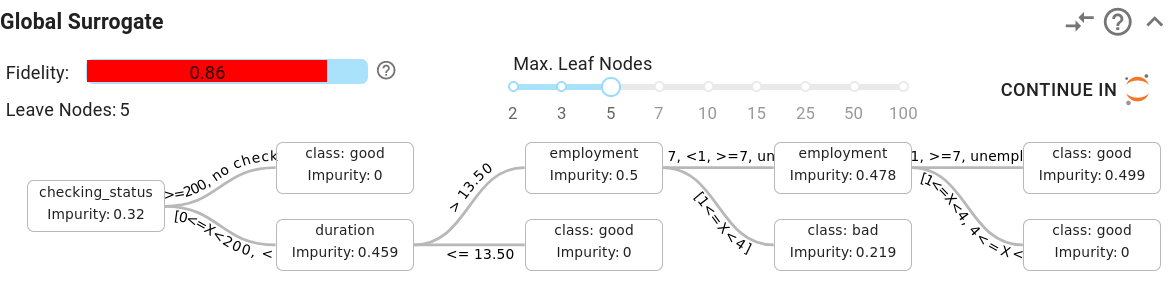}
        \caption{Global surrogate. A decision tree is rendered as a global surrogate. Using the top slider, the tree complexity can be controlled.}
        \label{fig:global_surrogate}
        \Description{Screenshot of global surrogate view. Fully described in the text.}
    \end{subfigure}
    
    \begin{subfigure}[b]{0.49\textwidth}
        \includegraphics[width=\textwidth]{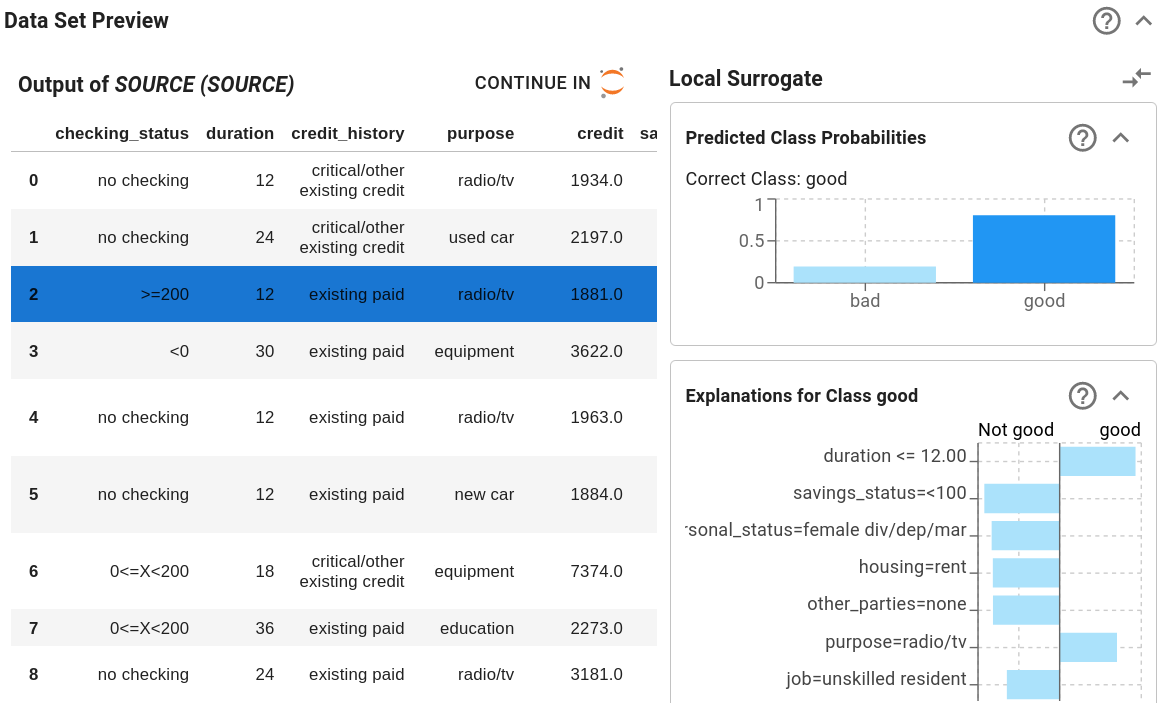}
        \caption{Dataset preview and local surrogate. The left-hand side displays a preview of the input dataset. On the right-hand side feature attributions for the selected sample are displayed.}
        \label{fig:data_set_preview}
        \Description{Screenshot of dataset preview view. On the left-hand side an extraction of the intermediate dataset is rendered as a table. On the right-hand side an optional assignment probability of the selected record to each potential class is displayed at the top. Down below the standard LIME visualization displaying the contribution of the various features to the predicted class is displayed.}
    \end{subfigure}
    \hfill
    \begin{subfigure}[b]{0.49\textwidth}
        \includegraphics[width=\textwidth]{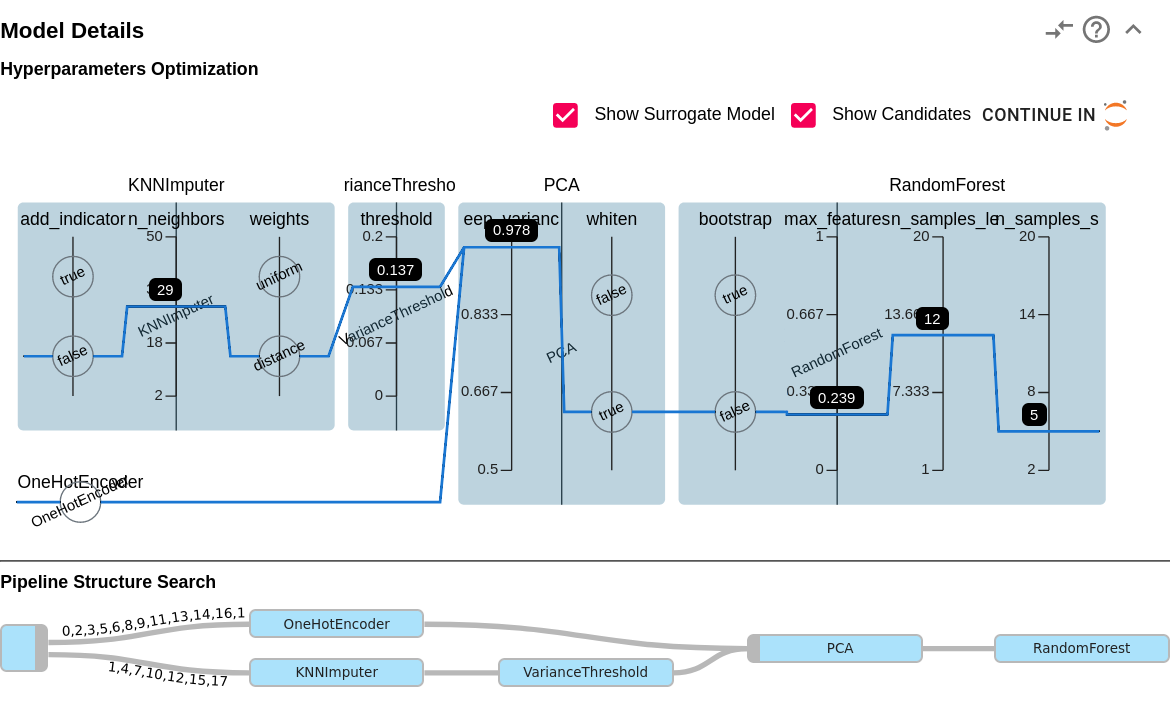}
        \caption{Model details. At the top, the selected hyperparameter configuration is displayed in a \ac{CPC}. At the bottom, the aggregated structure search graph up to this candidate is rendered.}
        \label{fig:candidate_configuration}
        \Description{Screenshot of the configuration view. Displayed are a conditional coordinate plot and pipeline structure search graph. Both visualizations are described in more detail in Figure \ref{fig:cpc} and \ref{fig:template_search_strategies}.}
    \end{subfigure}

    \begin{subfigure}[b]{0.49\textwidth}
        \includegraphics[width=\textwidth]{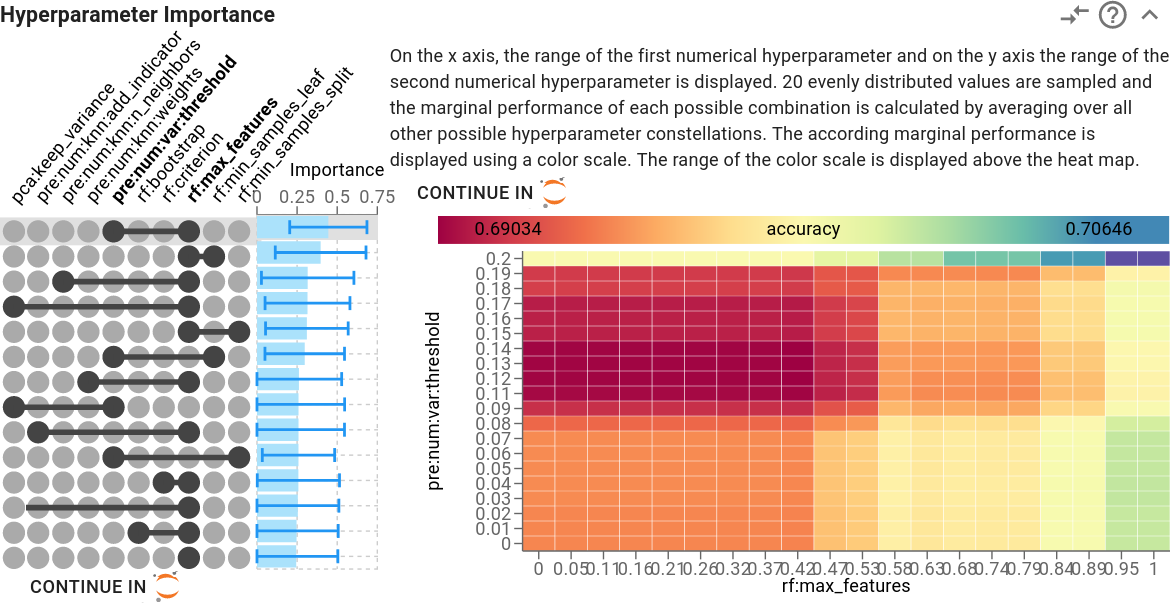}
        \caption{Hyperparameter importance with well-performing regions. On the left-hand side hyperparameter (pairs) are ranked by their importance. On the right-hand side a heat map with well-performing regions of the currently selected hyperparameter pair is displayed.}
        \label{fig:hp_importance}
        \Description{Screenshot of the hyperparameter importance view. On the left-hand side all hyperparameter (pairs) are listed sorted by their computed performance. The most important entry, the combination of two numerical hyperparameters, is selected. On the right-hand side, a heat map with the interactions of both selected hyperparameters is displayed. In addition, a description on how the read the heat map is given.}
    \end{subfigure}
    \hfill
    \begin{subfigure}[b]{0.49\textwidth}
        \includegraphics[width=\textwidth]{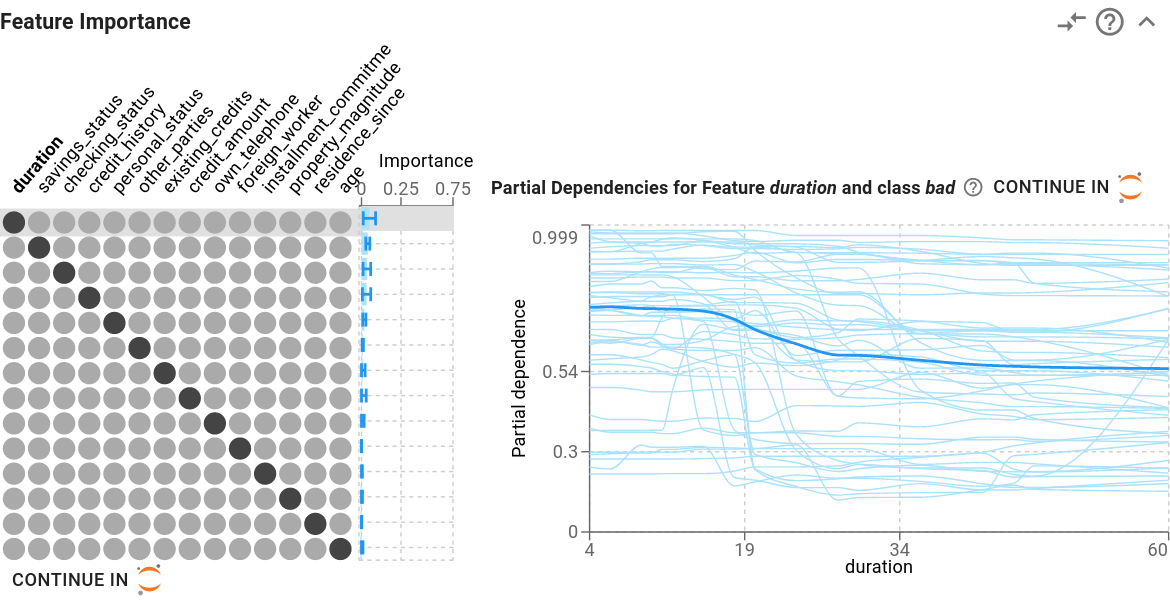}
        \caption{Feature importance with \ac{PDP} and \ac{ICE}. On the left-hand side the features are ranked by their importance. On the right-hand side a \ac{PDP} and \ac{ICE} plot show the correlation between the feature value and predicted class for the currently selected feature.}
        \label{fig:feature_importance}
        \Description{Screenshot of the feature importance view. On the left-hand side the same visualization as for the hyperparameter importance is used to list all features sorted by the importance. The most important feature is currently selected and the corresponding ICE and PDP plots on the right-hand side are given.}
    \end{subfigure}
    
    \caption{Overview of visualizations used to explain and validate a single candidate.}
    \label{fig:candidate_inspections}
\end{figure}

Users can reveal more details about candidates by selecting the corresponding entries in the leaderboard (Figure~\ref{fig:startpage}, C1). The \emph{candidate inspection} view provides data and model transparency for a single \ac{ML} pipeline. Users can find all required information to decide if the corresponding model is a ``good'' model worth selecting (N1). Therefore, existing \ac{XAI} techniques are incorporated to provide a holistic visual user interface. Users can interactively control the level of detail they are interested in. The requirements analysis revealed no clear preferences regarding the presented information. Therefore, we decided to implement most of the model and data inspections and provide measures for users to visualize or extract the missing information on their own.
Related information is aggregated in a \emph{card} with a short description (Figure~\ref{fig:startpage}, C2) that can be revealed by selecting it. Figure~\ref{fig:candidate_inspections} provides an overview of all implemented cards which are described in more detail later in this section. Inspections in each card are always computed based on the output of a single step in the pipeline. The complete pipeline structure is visible in a \emph{pipeline visualization} view (Figure~\ref{fig:startpage}, C4). It renders all pipeline steps as a \ac{DAG} that enables users to identify differences between the various candidates quickly (R13, R23). Even though the space for the pipeline visualization is quite limited, it is still sufficient to display typical pipelines constructed by modern \ac{AutoML} systems containing at most 10 nodes and only a few parallel paths, \eg, \cite{Feurer2015}. 

In Figure~\ref{fig:startpage}, a virtual data source node is currently selected and all inspections are calculated on the input data. Users can observe how the different steps in a pipeline affect the data by selecting the corresponding nodes in the pipeline visualization (Figure~\ref{fig:startpage}, C4) and inspecting the updated content of the different cards. The intention is to allow users to understand how each step in a pipeline modifies the input data. Each card provides at least one option to export the visualized data to \name{Jupyter}. Finally, users can also compare the information presented in a single card for various models (Figure~\ref{fig:startpage}, C5, R22, R23, R24).

The \emph{performance details} view (Figure~\ref{fig:performance_details}) provides basic performance metrics (R17) and visualizations (R18). Displayed are training and test performance, duration of the training, duration of predicting new samples, and an optional multi-fidelity budget \cite{Li2018} used for this model. The class report provides the precision and recall for each individual target class. In addition, a standard confusion matrix and \ac{ROC} curve are displayed. This view shall enable users to assess the performance of an \ac{ML} model in several dimensions. \name{AutoML} optimizes models against a single performance metric which may often be not enough to truly assess the predictive power and potential problems of a model. In addition, users can export the class report and confusion matrix to \name{Jupyter} for further analysis or visualization.

The \emph{global surrogate} view (Figure~\ref{fig:global_surrogate}) fits a decision tree to the predictions of the selected candidate (R19). Users can interactively control the size of the decision tree by specifying the maximum number of leaf nodes, effectively weighting the explainability of the surrogate versus the fidelity to the black-box model. The fidelity bar provides an estimate of how good the decision tree resembles the actual model. The idea is to provide an easy option to generate an interpretable surrogate model with adaptable complexity. This may be used to validate and explain the behaviour of an \ac{ML} model or even for legal auditions of it \cite{Mohseni2021}. Users can export the fitted decision tree to \name{Jupyter} for further analysis.

In the \emph{data set preview} (Figure~\ref{fig:data_set_preview}), users can inspect the output dataset of the currently selected pipeline step (R1, R5, R9). This allows users to observe how each step in the pipeline modifies the input data and provides data transparency. Even without detailed understanding of each pipeline step, users may be able to deduce the rough impact of a step on the data. We deliberately decided not to support data visualizations. Study participants stated highly varying goals and desires for visualizations, covering gaining data insights, viewing statistical distributions, and generating cohorts for the analysis of \ac{FATE}, just to name a few. Instead of providing only a limited data visualization---that would often be too restricted for users---, we rely on users to generate their own data visualizations in \name{Jupyter} and only provide tabular representations of the data. While this prevents users without knowledge in data visualization from visualizing the data, experienced users are not artificially limited by the visual analytics tool. Besides inspecting the tabular data, users can generate local surrogates (R20), computed via \name{LIME} \cite{Ribeiro2016}, for arbitrary records in the dataset. This provides a simple to understand attribution of feature values to the final prediction allowing users to gain insights into the model behaviour for single data instances to understand potential misclassifications.

The selected hyperparameters of each step in the pipeline are listed in the \emph{model details} view (Figure~\ref{fig:candidate_configuration}). The hyperparameters are displayed in a \ac{CPC} plot to visualize the respective search space (R14) and selected value for each hyperparameter (R21). In addition, the structure search graph at the time of sampling this pipeline structure is also displayed. More details regarding the \ac{CPC} and structure search graph are available in Section~\ref{sec:search_strategy}.

Hyperparameters can have a significant impact on the performance of an \ac{ML} model. In reality, only a few hyperparameters actually impact the performance significantly \cite{Hutter2014} making it important to identify these. The \emph{hyperparameter importance} view (Figure~\ref{fig:hp_importance}) visualizes the importance of single hyperparameters and interactions between pairs of them (N3). The hyperparameter importance is calculated using \name{fANOVA} \cite{Hutter2014}. By selecting a hyperparameter, users get a detailed overview of well- and bad-performing regions in the search space. Using these visualizations, experienced users can extract valuable insights on how to modify the search space for the next optimization round by removing hyperparameters or adapting the search space limits.

Finally, the \emph{feature importance} view (Figure~\ref{fig:feature_importance}) reveals information about relevant features. Using a permutation feature importance \cite{Breiman2001}, the impact of each feature on the predictive power of the \ac{ML} model is measured. Besides a ranking of all features, users can also view \ac{PDP} and \ac{ICE} plots. In case of a multi-class classification task, users can select the target class in \ac{PDP} and \ac{ICE} via a drop-down menu. This view should provide insights which features are important for the prediction and how feature values correlate with the predicted class. Both types of information can be used to validate the model behaviour.

\subsection{Inspecting the AutoML Optimization Procedure}
\label{sec:search_strategy}
Besides validating models produced by \ac{AutoML}, gaining insights to the actual \ac{AutoML} optimization procedure was ranked as a desired information during the requirements analysis. This process transparency can help users to understand what the search space looks like, how different pipelines are chosen, and which strategy is used to optimize hyperparameters (N2). Furthermore, users are given information helping them to decide if the search space was sufficiently explored and the optimization algorithm is converging to a local minimum.

\paragraph{Pipeline Structure Search}
A multitude of different search strategies have been proposed to synthesize pipeline structures. In general, pipeline structure search approaches can be divided into three different groups:
\begin{enumerate*}
    \item The simplest approaches use a \emph{fixed} pipeline structure that does not change during the optimization. Instead, only the hyperparameters of the individual pipeline steps are fine-tuned. This pipeline structure is usually hand-crafted based on best-practices.
    \item \emph{Template-based} approaches also utilise a best-practise pipeline, but single steps in the pipeline are restricted to a set of algorithms instead of a fixed algorithm. The optimizer is allowed to, for example, pick a classification algorithm on its own instead of only tuning an \ac{SVM}. 
    \item Approaches in the third group are able to build pipelines with \emph{flexible} shape. The pipeline shape can be adopted to the given problem instance freely based on some internal optimization strategy.
\end{enumerate*}

\begin{figure}
    \begin{subfigure}[b]{0.32\textwidth}
        \includegraphics[width=\textwidth]{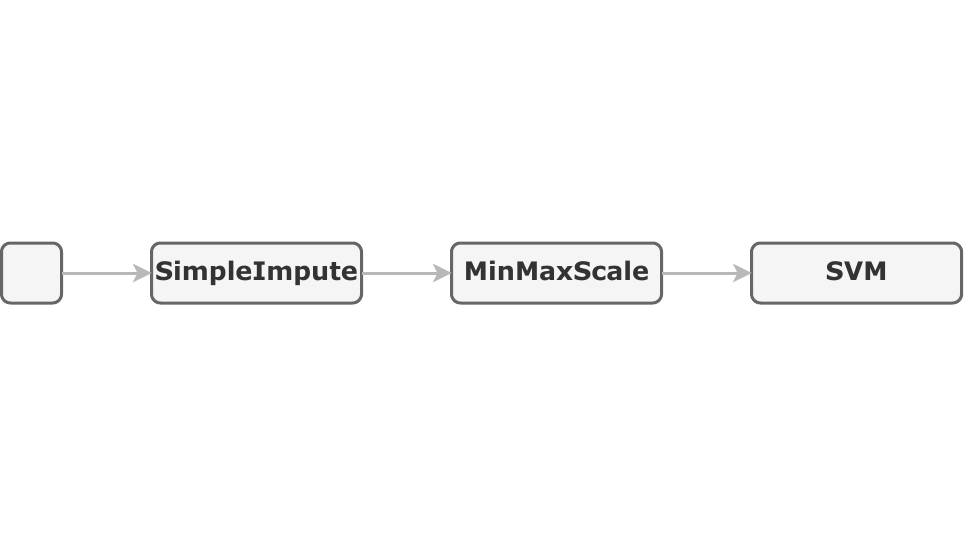}
        \caption{Iteration 1.}
    \end{subfigure}
    \hfill
    \begin{subfigure}[b]{0.32\textwidth}
        \includegraphics[width=\textwidth]{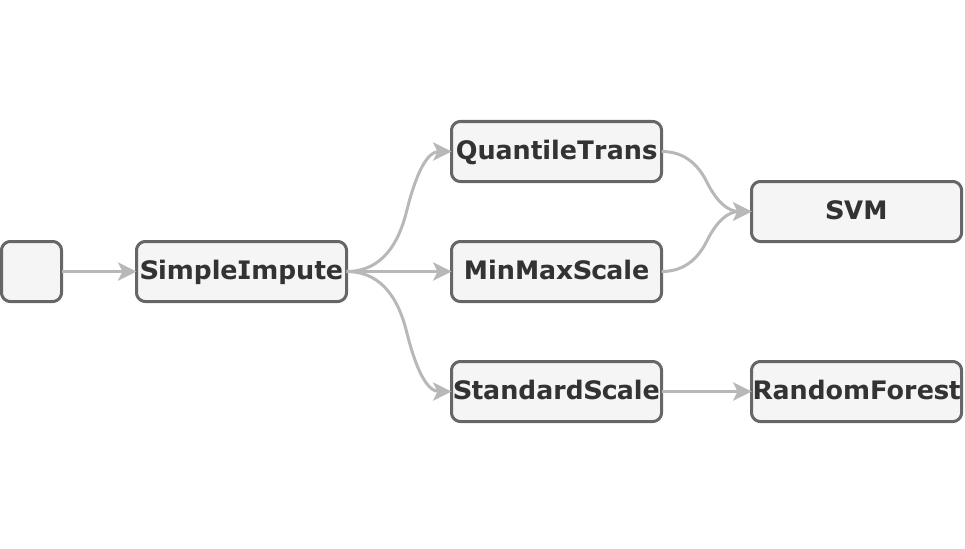}
        \caption{Iteration 3.}
    \end{subfigure}
    \hfill
    \begin{subfigure}[b]{0.32\textwidth}
        \includegraphics[width=\textwidth]{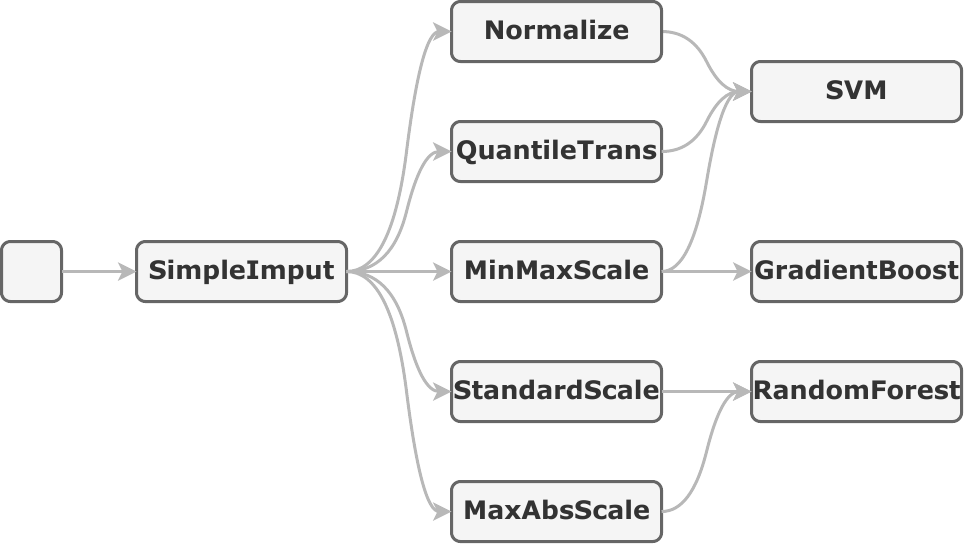}
        \caption{Iteration 6.}
    \end{subfigure}
    
    \caption{Example visualization of a template-based pipeline structure search strategy. The structure search graph is displayed after one, three, and six iterations. The search space contains pipelines with exactly three steps: First an imputation, followed by an arbitrary feature scaling, and finally an arbitrary classifier.}
    \label{fig:template_search_strategies}
    \Description{A template-based graph structure that becomes more complex with each iteration. In the first iteration, only a linear pipeline with three steps is displayed. After three iterations, the aggregated pipeline structure becomes more complex. The first pipeline step is still only a single algorithm, for the second step three alternatives can be selected, and for the last step two alternative are available. After the sixth iteration, the first step contains still only a single algorithm, but four and three different algorithms are available for the second and third step respectively.}

    \begin{subfigure}[b]{0.32\textwidth}
        \includegraphics[width=\textwidth]{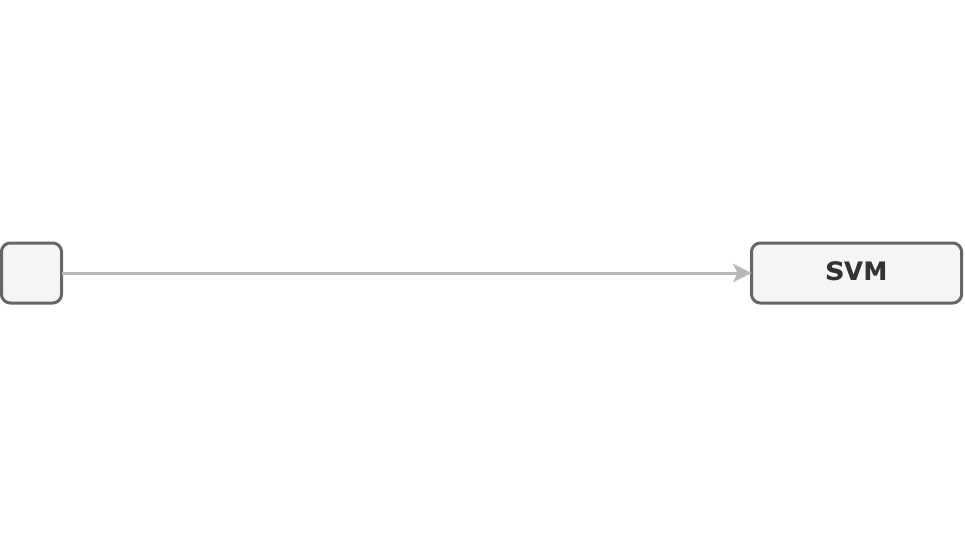}
        \caption{Iteration 1.}
    \end{subfigure}
    \hfill
    \begin{subfigure}[b]{0.32\textwidth}
        \includegraphics[width=\textwidth]{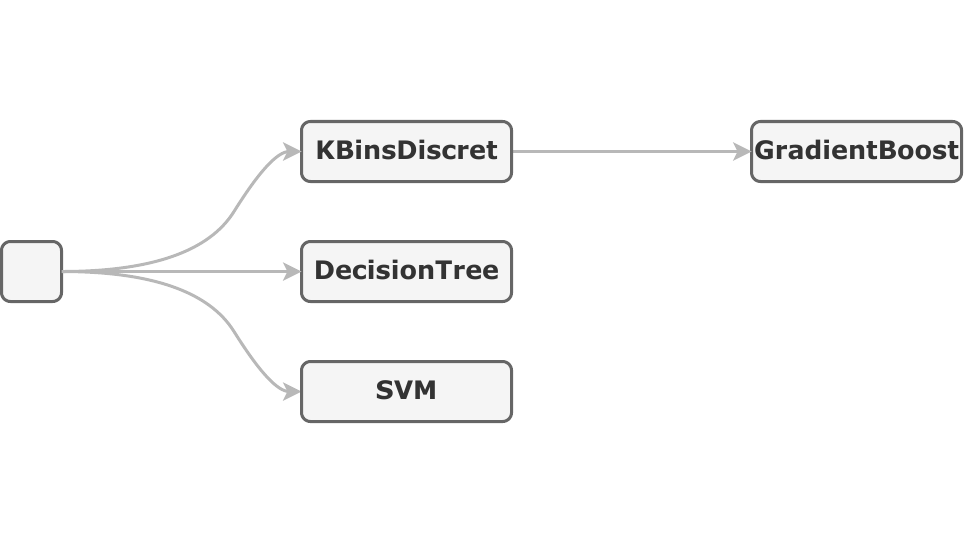}
        \caption{Iterations 3.}
    \end{subfigure}
    \hfill
    \begin{subfigure}[b]{0.32\textwidth}
        \includegraphics[width=\textwidth]{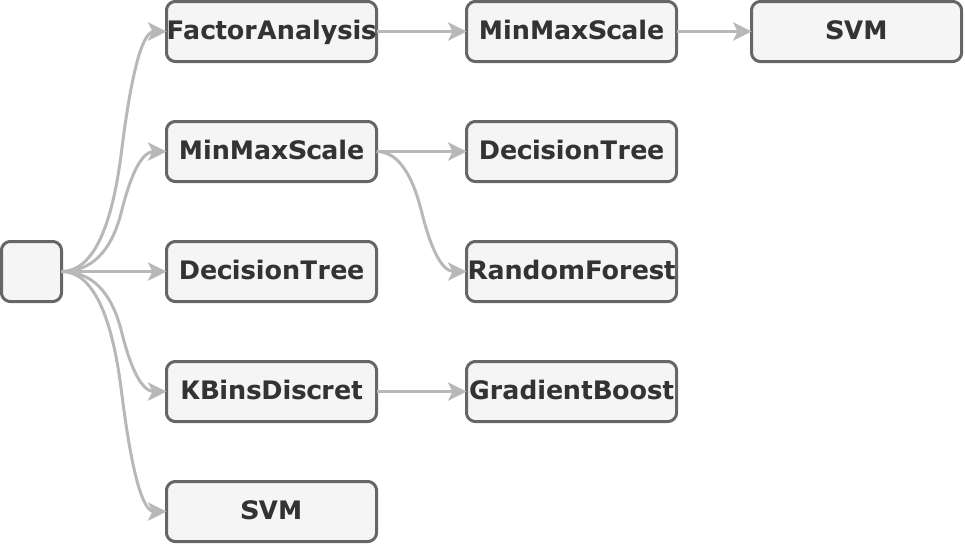}
        \caption{Iterations 6.}
    \end{subfigure}
    
    \caption{Example visualization of a flexible shaped search strategy. The structure search graph is displayed after one, three, and six iterations. The search space contains linear pipelines with different lengths and arbitrary pre-processing steps.}
    \label{fig:flexible_search_strategies}
    \Description{A flexible shaped graph structure that becomes more complex with each iteration. In the first iteration only a pipeline with a single step was produced. In the third iteration two pipelines with only a single step and one pipeline with two steps was produced. Finally, after six iterations, two pipelines with only one step, three pipelines with two steps, and a single pipeline with three steps were constructed.}
\end{figure}

For a user without strong \ac{AutoML} expertise, it is often not obvious what the pipeline structure search space actually looks like and how the \ac{AutoML} system traverses the search space. Consequently, users cannot easily deduce which kind of \ac{ML} pipelines they can expect to be constructed by an \ac{AutoML} optimizer. \name{XAutoML} provides a novel, intuitive visualization that allows users to grasp the general idea of the search procedure (R15), independent of the actual \ac{AutoML} system. Therefore, we iteratively merge all constructed pipelines into a single structure search graph. Using a time-lapse function, users can observe how this structure search graph is constructed iteratively and can deduce the approximate search strategy and the corresponding structure search space. Figures~\ref{fig:template_search_strategies}~and~\ref{fig:flexible_search_strategies} provide example visualizations for the template-based and flexible-shaped search strategies.

Usually, a single \ac{ML} pipeline is interpreted as a \ac{DAG} in which each node is an \ac{ML} primitive and edges indicate the flow of data between the primitives. \ac{ML} pipelines often contain parallel paths, \eg, to use different pre-processing steps for numerical and categorical features. In the merged graph, it is not directly apparent if two child nodes of a node are actually part of the same pipeline using parallel paths or two different pipelines using identical steps at the beginning of the pipelines. A distinction between these two cases is important to correctly assess the pattern of constructed pipelines. In the case of parallel paths, we provide a visual guidance by adding the names of selected columns to out-going edges.

Merging two pipeline structures to a joint structure search graph can be interpreted as a bipartite graph matching problem. To merge two \acp{DAG} \(G_1\) and \(G_2\), we use a simplified version of the graph matching algorithm proposed by \citet{Ono2021}: Similar to computing the edit distance between two strings using dynamic programming, a cost matrix \(|G_1| \times |G_2|\) with all possible substitutions, additions, and deletions is constructed. Using the Hungarian algorithm \cite{Kuhn1955}, a mapping with minimal cost of rows to columns is computed. Two nodes are considered identical if they substitute each other, \ie, their substitution is selected in the cost matrix. \(G_1\) and \(G_2\) are merged by creating a compound node for identical nodes identified in the previous step and adding all remaining nodes from \(G_1\) and \(G_2\) to the new graph. Starting with the first two sampled pipeline structures, we repeat this procedure iteratively to merge the next pipeline structure into the joint structure search graph up to the currently selected timestamp in the time-lapse.

\begin{figure}
    \centering
    \includegraphics[width=0.8\textwidth]{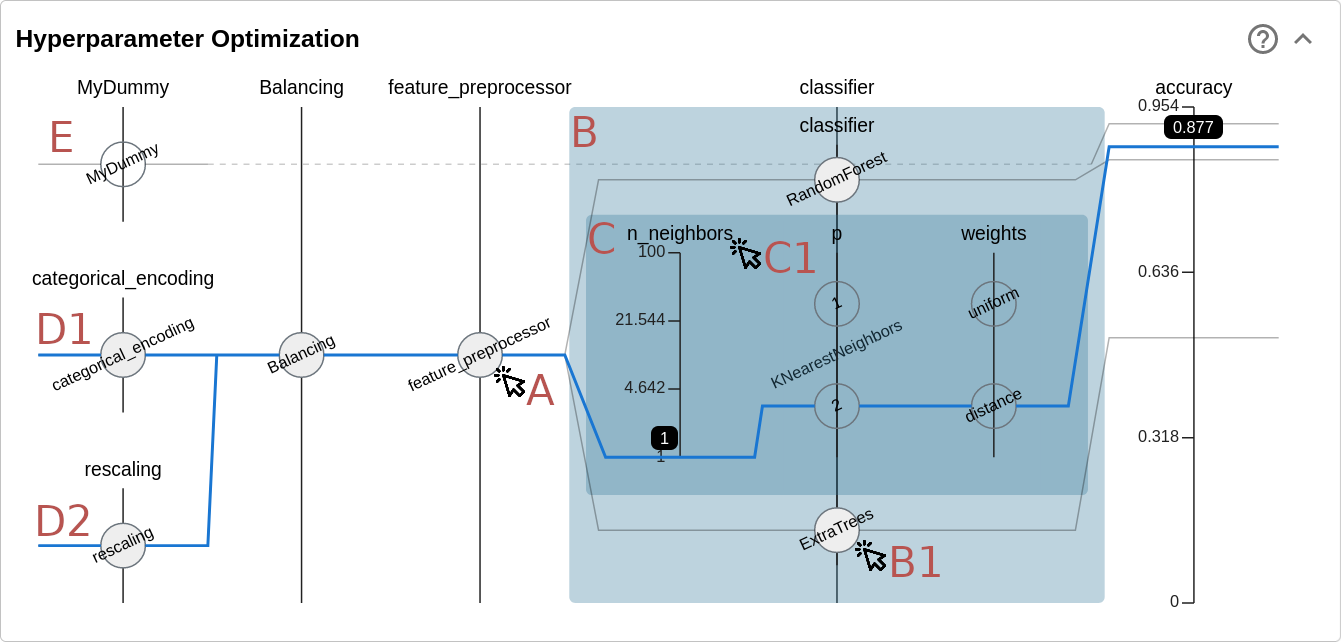}
    \caption{Extended \ac{CPC} visualization. Each axis at the top represents a single step in a pipeline. Each pipeline step can be expanded to reveal algorithms used in this step (A). In the example the classifier step is expanded (B, highlighted in light blue). Each algorithm in a step can be expanded again to reveal its hyperparameters (B1). In this example the \(k\)-Nearest Neighbors classifier is expanded (C, highlighted in dark blue). The classifier has three hyperparameters. Each of the inner axis contains the according search space with the actually selected values. Hyperparameters can be selected for the \emph{sampling history} view (C1). Each horizontal line represents a single candidate. One candidate with parallel steps is highlighted in blue (D1 and D2) resembling the actual pipeline structure. A second pipeline structure with empty values for multiple axes is displayed at the top(E).}
    \label{fig:cpc}
    \Description{A conditional coordinate plot including all improvements proposed in the text. The CPC show a pipeline with a parallel path and in total eleven different steps. Two steps are expanded to reveal to available algorithms in these steps. In addition, the k-nearest neighbours classifier is expanded to reveal its three hyperparameters. The CPC plot contains multiple candidates and a single candidate, passing through all expanded axes is highlighted in blue.}
\end{figure}

\paragraph{Conditional Parallel Coordinates for Non-Linear Pipelines}
The \acf{CPC} visualization proposed by \citet{Weidele2020} provides an intuitive way to inspect the hyperparameters of a complete \ac{ML} pipeline. It is an extension of the original parallel coordinates visualization \cite{Inselberg1990} that allows users to interactively control the amount of information shown. Instead of showing all hyperparameters at once, \ac{CPC} introduces different conditional layers of details. Users can drill down into particular steps of the \ac{ML} pipeline (see Figure~\ref{fig:cpc}, A), revealing more details, namely all potential algorithms that can be used in this step, in form of a parallel coordinate plot (see Figure~\ref{fig:cpc}, B). Each algorithm can be expanded again to reveal its hyperparameters (see Figure~\ref{fig:cpc}, B1). For each hyperparameter a parallel coordinate axis is plotted containing the complete search space of it (see Figure~\ref{fig:cpc}, C). This hierarchical stacking of the axes fits naturally with \ac{AutoML} search spaces. Search spaces are usually defined as tree structures due to conditional dependencies between hyperparameters \cite{Hutter2009}. To stick with the example in Figure~\ref{fig:cpc}, the pipeline highlighted in blue is configured to use a \(k\)-Nearest Neighbors classifier. Consequently, the hyperparameters of a random forest classifier are inactive as they do not change the behaviour of the pipeline.

The visualization proposed by \citet{Weidele2020} has two limitations, preventing the usage in combination with arbitrary \ac{AutoML} systems. First, \ac{CPC} requires a fixed number of coordinates present in all evaluated pipelines. This implies that all pipelines have to have the exact same number of steps. To resolve this limitation, we extend \ac{CPC} by introducing the option to not provide values for selected coordinates. These missing values are indicated by a dashed line (see Figure~\ref{fig:cpc}, E). Second, \ac{CPC} assumes a total order of pipeline steps. Yet, in reality, pipelines often contain parallel paths to use different pre-processing steps for categorical and numerical features. We extend \ac{CPC} to support non-linear pipelines by splitting the vertical space into multiple simultaneous coordinates (see Figure~\ref{fig:cpc}, D1 and D2) for steps containing parallel paths. Consequently, the axes in \ac{CPC} align with the actual pipeline shape again. This view should allow users to inspect the hyperparameter search space and sampled values of single hyperparameters for arbitrary complex pipeline structures.

\begin{figure}
    \begin{subfigure}[b]{0.45\textwidth}
        \includegraphics[width=\textwidth]{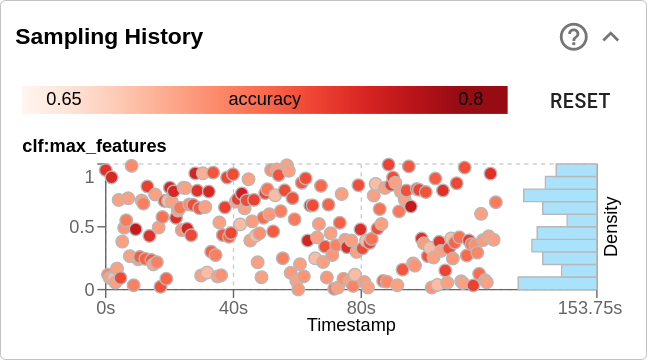}
        \caption{Random search.}
        \Description{A visualization of sampled hyperparameter values over time using random search. The scatter dots are quite evenly distributed over the complete search space. The density histogram on the right-hand side of the scatter plots is roughly uniformly distributed.}
    \end{subfigure}
    \hfill
    \begin{subfigure}[b]{0.45\textwidth}
        \includegraphics[width=\textwidth]{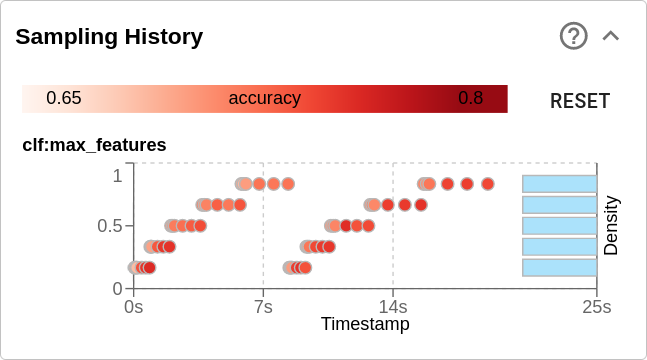}
        \caption{Grid search.}
        \Description{A visualization of sampled hyperparameter values over time using grid search. The scatter dots are perfectly evenly distributed over the complete search space. The density histogram on the right-hand side of the scatter plots shows a perfect uniform distribution of the sampled values.}
    \end{subfigure}
    
    \begin{subfigure}[b]{0.45\textwidth}
        \includegraphics[width=\textwidth]{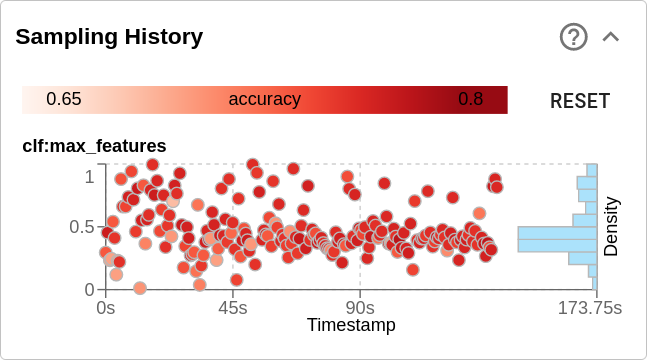}
        \caption{Bayesian optimization.}
        \Description{A visualization of sampled hyperparameter values over time using Bayesian optimization. At the beginning of the optimization, the scatter dots are randomly distributed over the complete search space. After roughly 30 seconds of optimization, more values are around a specific value in the search space. In addition, some values are still sampled at random. The density histogram on the right-hand side of the scatter plots roughly resembles a uniform distribution. Only around the specific value, much more samples were generated.}
    \end{subfigure}
    \hfill
    \begin{subfigure}[b]{0.45\textwidth}
        \includegraphics[width=\textwidth]{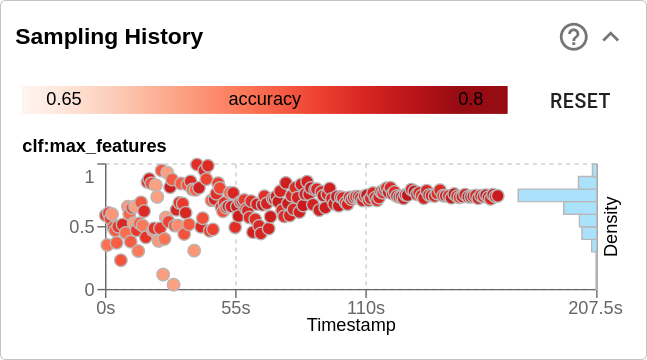}
        \caption{Population-based optimization.}
        \Description{A visualization of sampled hyperparameter values over time using population-based optimization. At the beginning of the optimization, the scatter dots are randomly distributed over the complete search space. After roughly 50 seconds of optimization, the sampled values start to converge to a specific value in the search space. Over time, the variance in the sampled values becomes smaller and smaller. The density histogram on the right-hand side of the scatter plots roughly resembles a Gaussian distribution.}
    \end{subfigure}

    \caption{Visualization of four different hyperparameter search strategies. Each scatter point represents a single \ac{ML} model constructed during the optimization. The \(x\)-axis represents the optimization duration; the \(y\)-axis contains the search range of the selected hyperparameter. The color coding is used to represent the performance of the individual models (better models have a darker color). On the right-hand side, a sampling distribution over the available search space is given as a histogram.}
    \label{fig:hp_search_strategy}
\end{figure}

The \ac{CPC} allows users to inspect single hyperparameters but the actual search strategy is not accessible as the sampling order of pipelines is not visible. To observe the sampling strategy of individual hyperparameters in a time-lapse (R16, N2), a \emph{sampling history} view---inspired by the visualizations in \name{HyperTendril} \cite{Park2021}---can be opened by selecting specific axes (see Figure~\ref{fig:cpc}, C1). The sampling history view, displayed in Figure~\ref{fig:hp_search_strategy}, shows the sampled values of single hyperparameters over time. Each individual hyperparameter is visualized using a scatter plot: Each scatter point represents a single \ac{ML} model, the \(x\)-axis presents the timestamp at which the model was sampled, and the \(y\)-axis represents the value of the selected hyperparameter. In addition, a color scale is used to encode the performance of each evaluated model such that users can confirm the optimization is converging to a local minimum. On the right-hand side, a histogram shows the distribution of sampled values over the possible search space. By selecting a scatter point, the details for the according model can be inspected. This view should allow users to deduce the rough hyperparameter search strategy. As an example, Figure~\ref{fig:hp_search_strategy} contains the visualization of four different search strategies that have very distinct sampling patterns. In addition, users can validate that indeed better models are found over time.

Finally, we integrate \ac{CPC} with the remaining user interface. Users can highlight individual candidates for a more detailed inspection by brushing numerical axes or selecting a choice in categorical axes. Further details for interesting candidates identified in \ac{CPC} can be accessed in the candidate inspection view by selecting the candidate.

\begin{figure}
\begin{minipage}{.5\textwidth}

\begin{algorithm}[H]
\caption{High-level description of the algorithm to visualize the exploration progress of the complete search space displayed in Figure~\ref{fig:optimization_progress}.}
\label{alg:config_distance}

\KwData{Evaluated Candidates \(\vec{\lambda}\), Performances \(\pi(\vec{\lambda})\), Hyperparameter Search Spaces \(\vec{\Lambda}\) }

Merge all search spaces \(\vec{\Lambda}\) into \(\Lambda_{merged}\)

Create artificial boundary candidates \(\vec{\lambda}_{boundary}\) based on \(\Lambda_{merged}\)

For each pair \(\lambda', \lambda'' \in \vec{\lambda} \cup \vec{\lambda}_{boundary} \), compute the distance \(d(\lambda', \lambda'')\) using Equation~\eqref{eq:config_distance}

Project \(\vec{\lambda}\) into a 2d-space using an \ac{MDS} based on \(d(\lambda', \lambda'')\)

Fit a regression model \(\hat{r}: \mathbb{R}^{2} \times \pi(\vec{\lambda}) \rightarrow \mathbb{R}\)

Create performance heat map using \(\hat{r}\)
\end{algorithm}

\end{minipage}
\hfill
\begin{minipage}{.45\textwidth}

\begin{figure}[H]
    \centering
    \includegraphics[width=\textwidth]{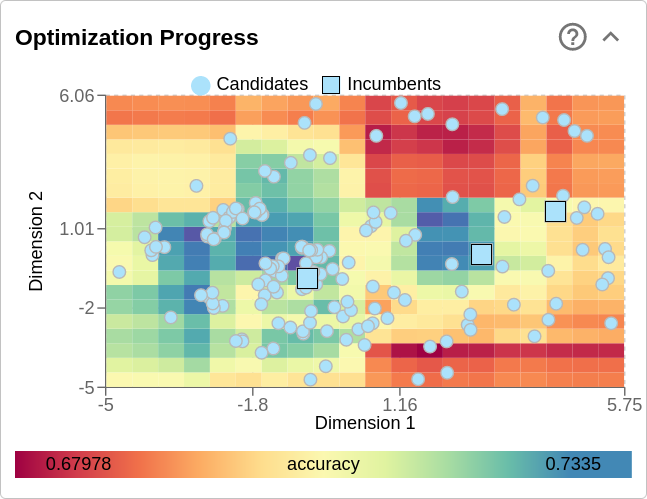}
    \caption{Visualization of the exploration and exploitation progress of the complete search space in two dimensions. Each scatter point represents a candidate, the heat map in the background provides an estimate of the performance.}
    \label{fig:optimization_progress}
    \Description{Visualization of the search progress using a scatter plot in combination with a heat map. The scatter dots are quite evenly distributed over the complete search space. At two points in the search space, clusters of scatter dots have formed. Those clusters are placed in regions which have a high accuracy according to the heat map indicating that the optimization has converged to a local minimum.}
\end{figure}

\end{minipage}
\end{figure}

\paragraph{Candidate Similarity and Search Space Coverage}
Current \ac{AutoML} systems do not provide measures for users to estimate the optimization progress. Users usually have to provide an optimization duration before starting the optimization but have no guidance on what duration is suited for their specific problem. An optimization duration is long enough if the search space was sufficiently explored and the optimizer converged to a local minimum. We propose a new procedure to visually inspect the coverage of an arbitrary \ac{AutoML} search space, based on a non-linear metric in combination with \ac{MDS} into two dimensions. A high-level overview of the procedure is given in Algorithm~\ref{alg:config_distance} and the final visualization in Figure~\ref{fig:optimization_progress}.

Given all previously evaluated candidates, their corresponding search space definitions, and their performances, we want to create a single plot that visualizes how well the search space was explored to provide guidance on a suitable optimization duration for users. As some \ac{AutoML} systems use a progressive widening of the search space, \eg, \cite{Zoller2021a}, a support of multiple search spaces is necessary. In the first step, we reuse the algorithm to merge different pipelines presented in the previous section to merge all hierarchical input search spaces into a single \ac{DAG} \(\Lambda_{merged}\). All existing candidates are padded with the default value of each newly introduced hyperparameter.

Next, boundary candidates \(\vec{\lambda}_{boundary} = \{\min(\lambda_1), \max(\lambda_1)\} \times \dots \times \{\min(\lambda_{|\lambda|}), \max(\lambda_{|\lambda|})\}\) are created with \(\min(\lambda_i)\) and \(\max(\lambda_i)\) being the lower and upper boundary of hyperparameter \(i\), respectively, and \(|\lambda|\) being the number of hyperparameters in \(\Lambda_{merged}\). For high-dimensional search spaces this step is skipped due to the combinatorial explosion of possible candidates. Next, the pair-wise distances between all candidates are calculated respecting the tree-structured search space. Therefore, we assume that the importance of a hyperparameter \(\lambda_i\) is directly dependent on its depth \(\delta(\lambda_i)\) in the hierarchical search space tree. The distance between two candidates is defined as a normalized heterogeneous euclidean distance weighted by \(\delta(.)\) as
\begin{equation}
\label{eq:config_distance}
    d(\lambda', \lambda'') = \sum_{i = 1}^{|\lambda|} dist \left( \lambda'_i, \lambda''_i \right).
\end{equation}
For a numerical hyperparameter \(\lambda_i\), \(dist(.)\) is defined as 
\begin{equation*}
    dist(\lambda'_i, \lambda''_i) = \left| \frac{\lambda'_i - \min(\lambda_i)}{ \left(\max(\lambda_i) - \min(\lambda_i)\right) \delta(\lambda_i)} - \frac{\lambda''_i - \min(\lambda_i)}{ \left(\max(\lambda_i) - \min(\lambda_i)\right) \delta(\lambda_i) } \right| ~.
\end{equation*}
In case of \(\lambda_i\) being a categorical hyperparameter, \(dist(.)\) is defined as 
\begin{equation*}
    dist(\lambda'_i, \lambda''_i) = \frac{1 - \mathbb{1}(\lambda'_i, \lambda''_i)}{\delta(\lambda_i)}
\end{equation*}
with \(\mathbb{1}\) being the indicator function.

Based on this distance matrix, all candidates are mapped into a 2d-space using an \ac{MDS}. Furthermore, a regression model \(\hat{r}: \mathbb{R}^{2} \times \pi(\vec{\lambda}) \rightarrow \mathbb{R}\) is fitted on the performance of the 2d candidates. This regression model is 
used to create a heat map showing the expected performance of the complete search space. In combination with an interactive time-lapse, users can check the sampled pipelines in the 2d representation over time. Virtually all modern \ac{AutoML} optimizers combine an iterative exploration of the search space with an exploitation of knowledge about well-performing regions. If the optimization duration was too short, only exploration was performed. Consequently, the evaluated candidates would be scattered uniformly over the complete 2d representation. Once well performing regions have been identified, clusters of similar pipelines begin to form in the 2d representation. If these well-performing regions stop changing and only new points are added to the already existing clusters, without creating new clusters, increasing the optimization duration will probably not lead to significant performance improvements. Observing the sampled values in the time-lapse, users should be able to deduce a suitable cutoff duration for the optimization.

\subsection{Ensemble Inspection}
Many \ac{AutoML} systems construct an ensemble of the best performing pipelines instead of yielding only the best candidate \cite{Feurer2015,Olson2016,Zoller2021a}. Therefore, \name{XAutoML} provides an \emph{ensemble inspection} view. This view contains the previously introduced candidate inspections for the complete ensemble. In addition, two visualizations are used to aid users in assessing the ensemble quality. We provide a tabular listing of all candidates in the ensemble with the according ensemble weighting. Users can open ensemble members in the candidate inspection view to get further details about the underlying \ac{ML} model. In addition, users can compare the individual predictions of all ensemble members on any record in the input dataset. 
Finally, we provide a decision surface plot for all ensemble members and the complete ensemble. A decision surface is a hyper-plane separating data points in the input data space according to their predicted class label. By mapping these hyper-planes into a two-dimensional space via a \ac{PCA} \cite{Pearson1901} and using a color-coding for regions separated by decision surfaces, the decision making of a classifier can be visualized as a heat map. Even though these decision surface plots only provide a very rough impression of the actual decision making of the models, users can still get an impression of how the different ensemble members behave. The decision surface plots shall enable users to quickly identify interesting ensemble members that behave differently from the rest. Those models can then be inspected in more detail.

\subsection{Dedicated Visualizations and Configuration}
\begin{figure}
    \includegraphics[width=\textwidth]{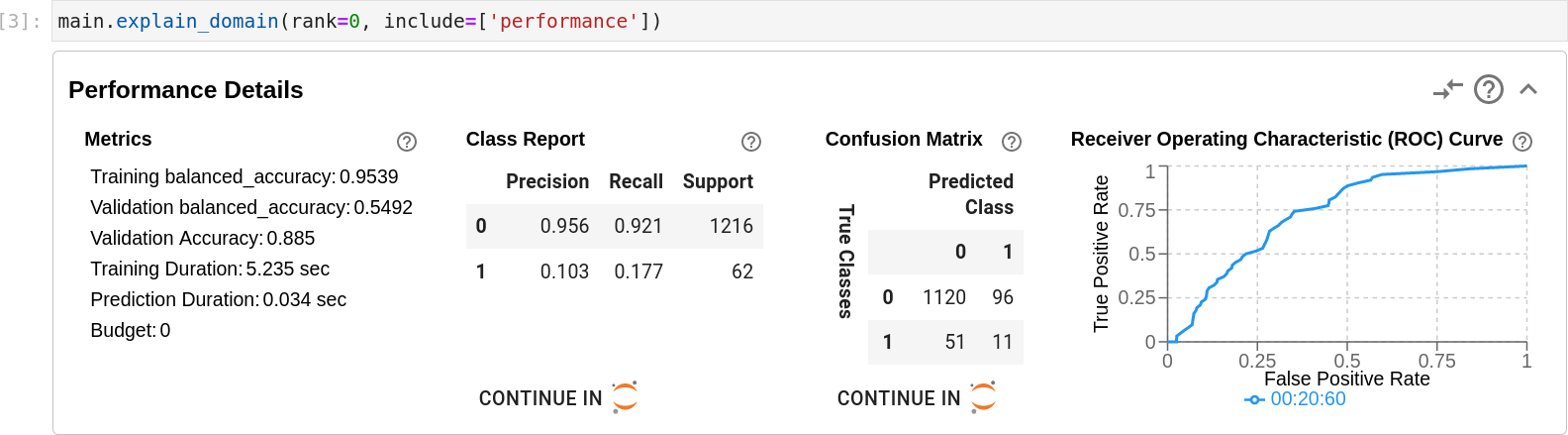}
    \caption{Dedicated performance details view of the best performing pipeline rendered in a separate \name{Jupyter} cell.}
    \Description{The performance details view described above is rendered in a dedicate \name{Jupyter} cell for the best performing pipeline candidate.}
    \label{fig:multiple_systems}
\end{figure}

While the visualization shown in Figure~\ref{fig:startpage} may provide a good opportunity for users to browse through all available information, it contains many superfluous information if a user is only interested in a specific subset of the available information. Therefore, \name{XAutoML} provides the option to render each of the visualizations presented in the previous sections in a dedicated \name{Jupyter} cell. Figure~\ref{fig:multiple_systems} shows an example of such a dedicated visualization.

These dedicated visualizations are intended to allow experienced users of \name{XAutoML} to create visualization tailored to their very specific visualization needs. Splitting the visualization into smaller chunks may allow users to weave \name{XAutoML} into the \textit{computational narrative} \cite{Rule2018} of their notebook. As all visualizations are still part of \name{XAutoML}, users only interact with different views into the same underlying system exposing them to a consistent look and feel.

Similarly, specific information can also be excluded from \name{XAutoML} via simple configurations. For example, a domain expert can configure \name{XAutoML} to only display the optimization overview, leaderboard, performance details, dataset preview and global surrogates. Consequently, users without expertise in \ac{ML} will not be exposed to technical \ac{ML} details while still allowing them to explore the complete optimization.

\subsection{Graphical Control Panel}
Originally, \name{XAutoML} was intended to be only used by users with programming expertise within a \name{Jupyter} notebook. However, this would basically prevent users without programming skills from using \name{XAutoML}. Therefore, we also provide a simple \emph{control panel} similar to \name{Visus} \cite{Santos2019}. This control panel allows users to upload a CSV file, select the target column from a tabular dataset preview and create an \ac{AutoML} optimization procedure with according configurations through a graphical user interface. For now, only a limited set of \ac{AutoML} frameworks with very basic configurations is supported. The optimization results are rendered in \name{XAutoML}, as described above, without the option to export insights as Python code to \name{Jupyter}.

\subsection{Implementation Details}
\name{XAutoML} is designed as a \name{JupyterLab} extension. The front-end is implemented in \name{React} \cite{React2013}, \name{Recharts} \cite{Recharts2015} and \name{D3} \cite{Bostock2011}. The back-end, responsible for everything related to model evaluations, is implemented as a \name{Python} package using \name{scikit-learn} \cite{Pedregosa2011}, \name{NumPy} \cite{VanDerWalt2011} and \name{pandas} \cite{McKinney2010}.

Execution results of \ac{AutoML} systems---namely the fitted models, a description of the complete search space, and some minimal meta-information---can be imported into \name{XAutoML} via a small wrapper code. This code is responsible for translating the \ac{AutoML} system specific formats to the one used by \name{XAutoML}. Developers of \ac{AutoML} systems can provide adapters for their systems with roughly 100 lines of code. Supported \ac{AutoML} systems are \name{dswizard} \cite{Zoller2021a}, \name{auto-sklearn} \cite{Feurer2015}, grid and random search from \name{scikit-learn} \cite{Pedregosa2011}, \name{FLAML} \cite{Wang2021b}, and \name{Optuna} \cite{Akiba2019}. More \ac{AutoML} systems may be added in future work.

To provide a seamless integration with \name{Jupyter} the imported execution results are rendered as a custom MIME type. The \name{Jupyter} extension connects this custom MIME type with the \name{React} application. In addition, the extension provides measures to export code snippets to new \name{Jupyter} cells which are used to extract artefacts from \name{XAutoML} and transfer them to \name{Jupyter}. Whenever the execution of \name{Python} code is necessary, \eg, for evaluating a single model, the extension provides a connection to the currently running user-kernel such that all code is executed in the same \name{Python} environment. The complete source code is publicly available on \name{Github}\footnote{
    Available at \url{https://github.com/Ennosigaeon/xautoml}.
} and can also be installed easily via the \name{Python} package manager.

\section{Evaluation}
\label{sec:evaluations}

In this section, we demonstrate the usefulness and effectiveness of \name{XAutoML} in a user study. As discussed by \citet{Muller2019}, many different roles are involved within data science projects. Similarly, \ac{AutoML} also does not aim to support a single user group but tries to assist multiple roles. Accordingly, an evaluation of \ac{AutoML} visualization methods should also be performed with multiple user groups. This user study was performed after the implementation of \name{XAutoML} was finished. We use the user study to answer the following three research questions:

\begin{description}
    \item[Q1] Does \name{XAutoML} provide \ac{ML}-related information understandable by a diverse user group?
    \item[Q2] Does \name{XAutoML} enable users with a diverse prior-knowledge of \ac{ML} and the underlying domain to validate \ac{ML} models?
    \item[Q3] Does \name{XAutoML} enable users with a diverse prior-knowledge of \ac{AutoML} to understand \ac{AutoML} optimizations?
\end{description}

\subsection{Study Protocol and Participants}
The user study was performed with the same 36 participants from the requirements analysis study in Section~\ref{sec:requirements-study} in a semi-structured method after developing \name{XAutoML}. To answer the three research questions, we compared a set of items (displayed in Table~\ref{tab:evaluation_results}) that participants were asked to answer for a baseline and for our proposed tool. These items aim to measure understanding of an \ac{AutoML} optimization and trust in the constructed \ac{ML} models. Identical input data was used to create the \ac{AutoML} optimizations that are visualized in the baseline and our approach. All participants answered the items for both the baseline and our tool in a fixed order. In addition, for the proposed tool participants should solve a set of information retrieval tasks to control if participants were able to understand the provided visualizations. Participants were invited to a roughly 60 minutes long online interview with one of the authors. During the session, participants shared their computer screens and sessions were recorded. Participants were encouraged to speak freely in a think-aloud study, as it was done in other evaluations of visual analytics tools, \eg, \cite{Wang2019,Weidele2020,Ono2021,Park2021}. We did not offer compensation for participating in the study. Each session consisted of four major phases:

\begin{enumerate}
    \item First, we welcomed the study participants and introduced them to the background and goals of our study. In addition, this phase was used to gain a better understanding of the participants abilities. Participants were asked to introduce themselves and their daily work in regards to data analysis, \ac{ML} and \ac{AutoML}. Next, participants were asked to gauge their proficiency in the three aforementioned knowledge dimensions domain expertise, \ac{ML} expertise and \ac{AutoML} expertise. Therefore, participants were asked if they agree with the statement \textit{``I am a domain expert''} on 4-point Likert scales with similar questions for \ac{ML} and \ac{AutoML}. Depending on their stated proficiency, participants were classified as either \textit{novices} (strongly disagree and disagree) or \textit{experts} (agree and strongly agree) in the respective area. In addition, participants without \ac{AutoML} expertise were given a short, high-level introduction to \ac{AutoML} from an end-user perspective to familiarise them with the goals and intended use cases of \ac{AutoML}.
    
    \item The second phase aimed at familiarizing the participants with the current state-of-the-art in \ac{AutoML}. They were provided a filled \name{Jupyter} notebook containing a typical \ac{AutoML} use-case\footnote{
        Example notebooks are available with the source code.
    }. First, basic information about the dataset was presented. The dataset was then used to train a classifier using \name{auto-sklearn}. Finally, the typical outputs produced by state-of-the-art \ac{AutoML} systems, \eg, \cite{Feurer2015,Olson2016,Wang2021b}, namely a textual representation of the best pipeline, the test loss of the best pipeline, as well as a visualization of the training loss during the optimization were provided. Participants were guided through the notebook and given time to inspect the source code in detail if they wished to. Based on this information, participants were asked to provide a first evaluation of their understanding of \ac{AutoML} to establish a baseline. Therefore, participants filled out a short questionnaire explained in more detail in Section~\ref{sec:main_insights} and displayed in Table~\ref{tab:evaluation_results}.
    
    \item Next, participants were asked to interact with \name{XAutoML}. To encourage the exploration of and to guide participants through the various aspects of \name{XAutoML}, participants were asked to solve small information retrieval tasks, which are described at the end of this section. A basic set of \(16\) tasks covering all major parts of \name{XAutoML} should be solved by all participants. In addition, more tasks were added if users were experts in a specific area (see Phase~1). The tasks were designed to be solved by different visualizations and therefore guide the participants through all parts of \name{XAutoML} introduced in Section~\ref{sec:xautoml}. The success or failure of solving each task was recorded by the interviewer. We reused the exact same input data from the second phase for all evaluations. After the participants had solved all assigned tasks, the understanding of \ac{AutoML} methods was evaluated again using the exact same questionnaire as in Phase~2 with the goal to compare the answers before and after using \name{XAutoML}.
    
    \item Afterwards, we wanted to test if the proposed option to create dedicated visualizations is useful for the participants. They were presented the options to configure \name{XAutoML} and render dedicated views. Using a 5-point Likert scale, participants were asked if this feature is helpful for them. Finally, the visual design and usability of \name{XAutoML} were recorded using the same 5-point Likert scale. In the end, participants were asked to reflect on any information that was missing in \name{XAutoML}. Therefore, participants were encouraged to examine the \name{Jupyter} notebook and \name{XAutoML} again. This opportunity was also used to let participants reflect on favored and unfavored elements of the proposed visual analytics tool. Finally, participants were given the chance to provide any remaining feedback freely.
\end{enumerate}

Participants with domain expertise were provided with a dataset tailored to their professional background, instead of participants using their personal ones, to eliminate effects of different datasets. They were selected to be
\begin{enumerate*}
    \item a classification problem, 
    \item a relevant problem in the corresponding domain,
    \item not trivially-solved by an \ac{ML} algorithm, and
    \item solvable by a domain expert.
\end{enumerate*}
For example, participants with a healthcare background were asked to create an \ac{ML} model predicting a high risk for a stroke based on 11 continuous and categorical features \cite{Fedesoriano2021}. Besides the raw data, participants were also provided with a description of all features and were allowed to familiarise themselves with the dataset. Participants without domain expertise were given one of the domain datasets at random.

The complete set of tasks participants solved in Phase~3 is listed in Table~\ref{tab:study_tasks} in the Appendix. They can be split into four groups:
\begin{enumerate*}
    \item \textit{General Information} regarding the optimization,
    \item \textit{Domain Expertise} for model validation and data insights,
    \item \textit{\ac{ML} Expertise} for model tuning, and
    \item \textit{\ac{AutoML} Expertise} for optimization understanding.
\end{enumerate*}
Each group contains tasks intended for all participants. In addition, a group may contain tasks requiring a profound prior knowledge in a knowledge dimension. Those tasks should only be solved by experts in the corresponding dimension.

\subsection{Results and Participant Feedback}
\label{sec:main_insights}
To answer our research questions, we analyze whether participants have a different understanding of the performed \ac{AutoML} optimization and the generated \ac{ML} models after using \name{XAutoML} in comparison to the baseline. We compared the answers to the questionnaire displayed in Table~\ref{tab:evaluation_results} that were given after using the baseline (Phase~2 of the study) with the answers after using the visual analytics tool (Phase~3 of the study). For each of the five questions in the questionnaire, the null hypothesis \(H_0\) is \textit{``There is no difference in the answers after using the baseline and after using \name{XAutoML}''} (\(H_0: \mu_{\mathrm{baseline}} = \mu_{\mathrm{XAutoML}}\)) for the current question under consideration. The alternative hypothesis \(H_1\) is that the usage of \name{XAutoML} influenced the responses (\(H_1: \mu_{\mathrm{baseline}} \neq \mu_{\mathrm{XAutoML}}\)). To compare the results, we performed two-sided \(t\)-Tests \cite{Gosset1908} with Holm-Bonferroni correction \cite{Holm1979}. We also provide effect sizes in terms of Cohen's \(d\)\cite{Cohen1988}.

\begin{table}
    \centering
    
    \caption{Summary of the questionnaire measuring the understanding of \ac{AutoML} and its generated models for domain experts (DE), data scientists (DS), \ac{AutoML} researcher (AR), and all together. For each question, the two rows represent results before and after using \name{XAutoML}, respectively. Values are given on a scale of 1 (strongly disagree) to 5 (strongly agree).}
    \label{tab:evaluation_results}

    \begin{tabularx}{\textwidth}{l X c c c c}
        \toprule
                             &                                                                                      & DE        & DS        & AR        & All \\
        \midrule
        \multirow{2}{*}{Q11} & \multirow{2}{=}{I understand how good the proposed model is.}                        & \(3.58\)	& \(3.27\)	& \(3.11\)	& \(3.23\) \\
                             &                                                                                      & \(4.53\)  & \(4.54\)  & \(4.44\)  & \(4.49\) \\
        \hline
        \multirow{2}{*}{Q12} & \multirow{2}{=}{I understand what the proposed model does.}                          & \(3.05\)	& \(2.77\)	& \(3.11\)	& \(3.03\) \\
                             &                                                                                      & \(4.32\)  & \(4.42\)  & \(4.11\)  & \(4.34\) \\
        \hline
        \multirow{2}{*}{Q13} & \multirow{2}{=}{I have enough information to decide if I want to use the model.}     & \(2.37\)	& \(2.54\)	& \(2.56\)	& \(2.43\) \\
                             &                                                                                      & \(4.21\)  & \(4.35\)  & \(4.22\)  & \(4.26\) \\
        \hline
        \multirow{2}{*}{Q14} & \multirow{2}{=}{I have an idea what happened during the \ac{AutoML} optimization.}   & \(2.84\)	& \(2.88\)	& \(3.00\)	& \(2.74\) \\
                             &                                                                                      & \(3.37\)  & \(4.19\)  & \(4.11\)  & \(3.71\) \\
        \hline
        \multirow{2}{*}{Q15} & \multirow{2}{=}{I would trust a model constructed by \ac{AutoML}.}                   & \(2.74\)	& \(2.85\)	& \(3.00\)	& \(2.77\) \\
                             &                                                                                      & \(3.89\)  & \(4.12\)  & \(4.22\)  & \(4.00\) \\
        \bottomrule
    \end{tabularx}
\end{table}

Overall, \name{XAutoML} was able to increase the trust in \ac{AutoML} systems significantly. On a scale from 1 (strongly disagree) to 5 (strongly agree), trust in \ac{ML} models constructed by \ac{AutoML} (Q15 in Table~\ref{tab:evaluation_results}) increased from \(2.77 \pm 0.94\) to \(4.00 \pm 0.80\), \(p < 0.0001\), \(d = 1.40\)\footnote{
    Given are mean \(\pm\) standard deviation, the corrected \(p\)-value, and the effect size in terms of Cohen's \(d\) \cite{Cohen1988}.
} after using \name{XAutoML}. Participants attributed the increased trust to a better understanding of the generated models and insights to the underlying optimization procedure. Accordingly, participants agreed to the statement that \name{XAutoML} is helpful for understanding and validating \ac{AutoML} (\(4.54 \pm 0.67\)). In the following, we take a closer look at how \name{XAutoML} improves process transparency and model validation. Furthermore, we discuss feedback provided by the participants. Details of all presented quantitative results are available in Tables~\ref{tab:evaluation_results} and \ref{tab:usability_results}.

\paragraph{Validating ML Models}
Participants used \name{XAutoML} extensively to validate and compare \ac{ML} models. Depending on their background in \ac{ML}, different explanations were utilised. Unsurprisingly, users without knowledge in \ac{ML} mainly focused on the performance metrics and global surrogates to gain insights into the behaviour of models. By inspecting multiple performance metrics on the train and test data in combination with the confusion matrix, users stated they received enough information to correctly assess the performance of a model (Q11, \(3.23 \pm 1.06\) versus \(4.49 \pm 0.56\), \(p < 0.0001\), \(d = 1.47\)). After picking a promising model, those users usually investigated the global surrogate to check if the model ``does something sensible'' (P23). Local surrogates and feature importance were consulted less often.

Experienced \ac{ML} users did not focus on the global and local surrogates and feature importance. This can at least partially be attributed to the analysed dataset being selected at random. Consequently, those explanations had only minimal potential to provide new insights. Instead, participants focused more on gaining insights by inspecting the used algorithms in the different pipelines. \ac{ML} practitioners liked the visualization of pipeline structures. The ability to observe changes in the data and the possibility to extract each intermediate dataset were considered very useful both for ``visualizing the data'' (P3) as well as ``pipeline debugging'' (P5). Some participants were able to detect ineffective or superfluous steps in the constructed pipelines using the intermediate datasets. Consequently, all participants were able to understand the behaviour of the constructed \ac{ML} models significantly better using \name{XAutoML} (Q12, \(3.03 \pm 1.22\) versus \(4.34 \pm 0.73\), \(p < 0.0001\), \(d = 1.31\)) confirming research question Q2.

\paragraph{Understanding the AutoML Optimization Procedure}
The second analytical need supported by \name{XAutoML} is aiding users in understanding the underlying \ac{AutoML} search strategy and search space. Even though \name{XAutoML} was able to increase the understanding of the \ac{AutoML} system (Q14, \(2.74 \pm 1.31\) versus \(3.71 \pm 1.23\), \(p = 0.0011\), \(d = 0.76\)), the results highly depend on the participant's prior experience with \ac{ML}. Users with little to no knowledge in \ac{ML} had gained only minor insights into the \ac{AutoML} procedures. They usually stated either missing knowledge in \ac{ML} or math to understand the internals or were simply not interested in it.

In contrast, data scientists and \name{AutoML} researchers were able to gain insights into the \ac{AutoML} optimization procedure. Even participants with no prior practical experience with \ac{AutoML} were able to correctly identify and distinguish template-based, fixed, and flexible optimization strategies for pipeline structures. ``The visualization is a good starting point to get into \ac{AutoML}'' (P32). Multiple users were surprised by the complexity of \ac{ML} pipelines produced by some \ac{AutoML} systems. ``I thought \ac{AutoML} is only good for hyperparameter optimization and pipeline construction is just a small extension'' (P10).

Similarly, most users were able to correctly identify different hyperparameter search strategies like random search, grid search, or a model-based optimization. ``Observing the search in a time-lapse is extremely helpful to understand the search algorithms'' (P9). The combination of structure search graph and \ac{CPC} enabled users to quickly identify the hyperparameters for each step in a potential \ac{ML} pipeline. Nearly all users were able to formulate a rough concept of the underlying search space confirming research question Q3.

P12, an experienced \ac{AutoML} user, highlighted the usefulness of the progress view for assessing the progress of the optimization run. ``It is often quite hard to judge if the optimization has actually converged yet. The overview gives me this information at a glance'' (P12). Yet, they also remarked that providing this information in real-time during the optimization procedure would be even more helpful. We discuss this issue of human guidance in Section~\ref{sec:discussion}.

\paragraph{AutoML in the Data Science Workflow}
The integration with \name{Jupyter} and the option to export datasets and \ac{ML} artifacts from \name{XAutoML} to \name{Jupyter} was received overwhelmingly positive. Only a single participant with programming skills said this integration is just a ``nice gimmick'' (P8) while all other participants highlighted the usefulness of this feature for their workflow. ``Visualizations can only offer a subset of all potential analyses. Experts with domain knowledge are able to extract more information'' (P2). Many participants confirmed our intended workflow presented in Section~\ref{sec:human-in-the-loop}. They stated they could imagine using \ac{AutoML} for creating an evaluation baseline in a new project they would aim to improve. Visual analytics are considered an integral part of \ac{AutoML}. ``Just because I was able to validate results in a single case, I will not trust \ac{AutoML} in general'' (P1). The current integration with \name{Jupyter} and the export of various artefacts from \name{XAutoML} does not force users to adapt their usual workflow. ``There is simply no reason not to use the visualization'' (P14). Unsurprisingly, the \name{Jupyter} integration was ignored by users without programming skills.

\paragraph{Grounding the Expectation in ML}
An interesting effect we were able to observe is that participants' proficiency in \ac{ML} correlates negatively (\(r = -0.33\)) with the understanding of what the proposed \ac{ML} model does. Without providing additional explanations, participants with a low proficiency in \ac{ML} reiterated the goal of the scenario when asked if they understood what the model does instead of questioning the actual behaviour of the \ac{ML} model. Even when challenging this assumption, participants were still convinced that the provided \ac{ML} model would solve their task. This is a good example of the risk of potentially ``automating bad decisions'' \cite{Crisan2021} without noticing it by practitioners with minimal \ac{ML} expertise. During the usage of \name{XAutoML}, most domain experts discarded the model initially suggested by the \ac{AutoML} optimizer quickly using the global surrogate and performance metrics. Consequently, domain experts were able to identify invalid models, effectively preventing this pitfall.

\paragraph{Validating the Behaviour of the Search Algorithm}
P11, an \ac{AutoML} researcher, tested \name{XAutoML} with an \ac{AutoML} system building increasingly more complex pipelines over time. The system uses an \(\epsilon\)-greedy approach to explore new pipelines in the beginning of the optimization and picks from the already evaluated pipelines toward the optimization end. Using the time-lapse function of the structure search graph, P11 observed how during the first half of the optimization mostly new pipelines with increasing length are sampled, just as they expected. Yet, toward the end of the optimization, the search algorithm suddenly started favouring very short pipelines. Longer pipelines that were also previously evaluated are completely ignored. Confused by this observation, P11 checked the implementation of the search procedure after the user study and detected a bug enforcing the selection of the shortest possible pipeline in an exploitation step. Even though this information would have also been available by carefully studying the evaluated pipelines in the log file, \name{XAutoML} revealed this flaw in under two minutes.

\paragraph{Information Overload}
In general, users were satisfied with the information presented in \name{XAutoML} as they had enough information to decide if they want to use a model constructed by \name{AutoML} (Q13, \(2.43 \pm 0.70\) versus \(4.26 \pm 0.61\), \(p < 0.0001\), \(d = 2.79\)). Even though not all information were relevant for all users, participants acknowledged the benefit of showing too much information instead of too little. ``Users differ very much in their experience and knowledge. I like that people are able to pick the information they need'' (P23). Furthermore, the information were presented in an understandable way as participants were able to solve \(96.2\%\) of the given tasks with no big differences between tasks for all users (\(96.56\%\)) and expert-only tasks (\(95.33\%\)). However, the current way of providing the information in \name{XAutoML} can still be improved. Multiple participants, with diverse proficiencies, remarked a visual information overload \cite{Poursabzi-Sangdeh2021} during the first minutes of the user study. ``There are too many features to grasp the complete program immediately'' (P3). Yet, on average, participants agreed to the statements that the visual design is easy to learn (\(4.09 \pm 0.77\)) and understand (\(4.03 \pm 0.74\)). From these apparently contradicting statements, we deduce that a better structuring of the presented information is necessary. Consequently, the research question Q1 is only partially fulfilled.

\begin{table}
    \centering
    
    \caption{Summary of the questionnaire measuring the visual design and usability of \name{XAutoML} for domain experts (DE), data scientists (DS), \ac{AutoML} researcher (AR), and all together. Values are given on a scale of 1 (strongly disagree) to 5 (strongly agree).}
    \label{tab:usability_results}

    \begin{tabularx}{\textwidth}{l X c c c c}
        \toprule
            &                                                                   & DE        & DS        & AR        & All      \\
        \midrule
        Q21 & The visual design is easy to learn.                               & \(4.05\)	& \(4.19\)	& \(4.22\)	& \(4.09\) \\
        Q22 & The visual design is easy to understand.                          & \(4.11\)	& \(4.04\)	& \(3.89\)	& \(4.03\) \\
        Q23 & \name{XAutoML} provides sufficient details for me.                & \(4.42\)	& \(4.42\)	& \(4.22\)	& \(4.37\) \\
        Q24 & \name{XAutoML} helps me to understand and validate \ac{AutoML}.   & \(4.37\)	& \(4.69\)	& \(4.67\)	& \(4.54\) \\
        Q25 & \name{XAutoML} helps me with my computational narrative.          & \(3.42\)	& \(4.38\)	& \(4.11\)	& \(3.89\) \\
        Q26 & The configuration of \name{XAutoML} is helpful for me.            & \(3.68\)	& \(3.58\)	& \(3.78\)	& \(3.71\) \\
        \bottomrule
    \end{tabularx}
\end{table}

\paragraph{Dedicated Views and Configuration}
The option to render dedicated views of \name{XAutoML} received mixed ratings from the participants. Some data scientists highlighted the usefulness for presentations (P27, P30) or training (P31) while others ignored it. Most \ac{AutoML} researchers and domain experts simply did not intend to use \name{Jupyter} for documentation or narrative purposes. Yet, as this feature is completely optional, on average, participants had a neutral feeling (\(3.89 \pm 1.29\)) about it. Similarly, participants had mixed feelings about the usefulness of configuring the displayed information in \name{XAutoML} (\(3.71 \pm 1.13\)). While participants being proficient in only a single knowledge dimension acknowledged the benefits from hiding irrelevant information, other participants were indifferent to this option. In contrast, some participants even disliked it because they want to browse through all available information.

\subsection{System Usability}
The usability of \name{XAutoML} is evaluated using the \ac{SUS} \cite{Brooke1996}. To compute the \ac{SUS}, participants were asked to fill out the standard \ac{SUS} survey at the end of the user study. The \ac{SUS} contains 10 questions participants were asked grade on a scale of \(1\) (strongly disagree) to \(5\) (strongly agree) and ranks a system on a scale from 0 to 100. Our system reached \(77.5 \pm 11.07\) points with detailed results being available in Table~\ref{tab:sus} in the Appendix. According to \citet{Bangor2008}, this grades the usability of \name{XAutoML} in the middle region of Good making it acceptable. Domain experts ranked the system worse (\(76.32 \pm 11.28\)) than data scientists (\(79.81 \pm 10.88\)) or \name{AutoML} researchers (\(78.89 \pm 11.87\)). The major criticism of participants was that they had problems imagining most people would learn to use the system quickly (\(7.14\) of \(10\) possible points). We attribute this assessment to the information overload experienced by participants.

\section{Discussion \& Future Work}
\label{sec:discussion}

Study participants were interested in validating the behaviour of single \ac{ML} models and/or understanding the internals of the \ac{AutoML} optimization. This confirms our identified visualization needs N1 and N2. It is important to note that these two needs do not correlate with the traditional user groups of data scientists and domain experts but depend on a more diverse personal background knowledge of each participant.

Interestingly, most participants with \ac{ML} expertise considered \ac{AutoML} only as a baseline method for \ac{ML}. Instead of using a proposed model directly, they wanted to build a custom model manually improving on the performance yielded by \ac{AutoML}. A potential explanation could be that most study participants had no prior experience with \ac{AutoML} and were therefore satisfied with \ac{AutoML} building ``good enough models'' (P3) for a baseline. Finding the best possible \ac{ML} model via \ac{AutoML} was not a goal of the participants. Consequently, a search space refinement---and therefore identifying well-performing regions in the search space---for consecutive optimization runs was not interesting for the study participants. This contradicts our identified visualization need N3. Further research with users having a wider spectrum of prior knowledge in \ac{AutoML} is necessary to validate our identified analytical needs.

Instead of designing \name{XAutoML} for a single user group, we decided to provide a lot of information and rely on users to pick the relevant and ignore the rest. The user study showed that participants from all three knowledge dimensions were able to extract relevant information from \name{XAutoML} supporting our design goal G1. Furthermore, study participants having programming skills highlighted the benefits of integrating \name{XAutoML} with \name{Jupyter}. Extracting data from the visual analytics for further manual analysis was considered an integral part for incorporating \ac{AutoML} and the visual analytics into the usual workflow of the participants confirming design goals G2 and G3. Finally, \ac{AutoML} systems can be integrated with \name{XAutoML} easily as only minimal information has to be provided by the \ac{AutoML} system supporting our design goal G4. To prove this claim, \name{XAutoML} is currently able to visualize the results of five \ac{AutoML} libraries.

As an alternative to using only a single visualization system, it would also have been possible to create multiple systems specialized for specific user groups, \ie, users with identical knowledge dimensions. We consider this approach suboptimal for several reasons:
\begin{enumerate*}
    \item A dedicated visualization for each knowledge dimension would expose users to up to three different systems. From a users perspective it is cumbersome to switch between different systems to collect information about basically one object of interest.
    \item The requirements engineering revealed that even within only a single user group, \eg, pure domain experts, a clear preference which information should be available is not visible. This would expose users to undesired information even with multiple systems.
    \item Some information can only be extracted when considering multiple knowledge dimensions at once, \eg, the impact of different preprocessing strategies on certain features related to \ac{FATE}. This combination of insights would be hard to achieve with distinct systems.
\end{enumerate*}
Even though \textit{one size fits all} visualizations are extremely difficult to design in the context of \ac{AutoML}, we still argue that a single system with the option to hide undesired information is the best compromise between overwhelming users with too many information and limiting users by not providing relevant information.

\subsection{Current Limitations}
Many participants stated that they were overwhelmed by the amount of different information presented in the visualization. This was the major criticism of \name{XAutoML}. We tried to prevent this visual information overload by designing \name{XAutoML} to provide information in different levels of details supporting our design goal G3. Yet, apparently, the current visual design fails at this task at least partially. This initial shock was usually overcome within roughly \(40\) minutes of participants interacting with \name{XAutoML}. Further analysis on how to restructure the presented information and visualizations is necessary. During the study several participants suggested introducing additional layers to expose users to less information at once. Alternatively, it would be interesting to let users interact with pre-defined configurations of \name{XAutoML} tailored to their knowledge dimensions to show less information at once.

\name{XAutoML} is currently designed to create post-hoc explanations. Users can only analyse the results of an \ac{AutoML} optimization run after it has finished. Based on the extracted insights, users can adapt the search space and dataset to guide the next optimization run into the right direction. From a user's perspective, this workflow can be cumbersome as the lengthy completion of the \ac{AutoML} optimization has to be awaited. In the spirit of human-guided \ac{ML}, a tight coupling between \ac{AutoML} system and visual analytics is highly desirable \cite{Gil2019a} to control the optimization procedure and modify the search space in-place during the optimization. Yet, this directly conflicts with the design goal to support many \ac{AutoML} systems. In addition, it exposes users to internal information of the \ac{AutoML} optimizer that could be hard to understand for end-users. On the other hand, multiple \ac{AutoML} researchers requested exactly these internals for detailed system debugging. Further research is required to distill common user interactions with \ac{AutoML} systems to a standardized framework coupling visual analytics and optimizer.

As a consequence of the post-hoc explanations and the abstraction of the underlying \ac{AutoML} system, most visualizations are computed on the fly. These ad-hoc computations become quite time-consuming for large datasets, severely limiting the user experience. Even though all computed results are cached and the input data for costly operations is down-sampled, the initial computation of the requested data may still require too much time for an interactive system. Another drawback of being \ac{AutoML} systems agnostic is the lack of information about the internal optimizer reasoning. \name{XAutoML} only provides visualizations based on the evaluated models over time. Given this information, users of \name{XAutoML} are able to make an educated guess about the internal reasoning of the \ac{AutoML} optimizer. It would be better if \ac{AutoML} systems provide more inherent information to explain their behaviour, but unfortunately, this is simply not the case with current state-of-the-art \ac{AutoML} systems.

Multiple participants remarked that relevant information, they were looking for, was not easy to find due to a discrepancy in the used terminology. \name{XAutoML} uses established terms from \ac{ML}, yet participants are often not familiar with \ac{ML} and use other terms for the same concept, \eg, \emph{recall} for binary classification and \emph{sensitivity}. The effort to mentally map to known terms made usage of \name{XAutoML} unnecessary straining for some participants.

During the study, multiple study participants were interested in filtering the displayed \ac{ML} candidates by various factors for further analysis. For example, participants wanted to use the importance of specific features for the analysis of \ac{FATE} or the confidence of model predictions. Currently, it is not possible to filter candidates at all to ease the analysis.

In the user study, we compared \name{XAutoML} with a pure programmatic method of using \ac{AutoML} in \name{Jupyter}. A comparison with an existing visual analytics tool as a baseline was not performed. Consequently, the observed improvements could potentially be attributed to the pure existence of (simple) visualizations and not the concrete design of \name{XAutoML}. Further insights, if similar improvements could be achieved with simpler approaches would be helpful to steer the future development of visual analytics tools for \ac{AutoML}.

\subsection{Future Work}

During the user study, many users voiced their desire to create specific \ac{ML} models based on the insights extracted from their detailed analysis of \ac{ML} models, \eg, giving a higher importance to a specific feature. \citet{Gil2019a} have analysed common requirements for human-guided \ac{ML}. An option to incorporate human guidance into a running \ac{AutoML} optimization, building single models specifically requested by the user, would be useful.

Currently, \name{XAutoML} provides the same visualizations for all \ac{ML} pipelines. Yet, some \ac{XAI} techniques are only applicable to specific \ac{ML} models, \eg, attention mechanisms for neural networks. By considering the actual \ac{ML} model of a pipeline, it would be possible to include additional model-specific visualizations.

\name{XAutoML} is currently limited to classification tasks on tabular data. An extension to further learning tasks and data types is desirable and was requested during the user study by multiple participants. Similarly, we plan the inclusion of further \ac{AutoML} systems to increase the potential user basis of \name{XAutoML}.

Finally, we only considered the interaction of a single user with \name{XAutoML}. However, in practice data science tasks are often solved by cross-functional teams. It would be interesting to evaluate in further studies how \name{XAutoML} could be used to assist multiple users.

\section{Conclusions}

We presented \name{XAutoML}, an open-source visual analytics tool for understanding and validating \ac{AutoML} procedures by making \ac{AutoML} optimizations transparent and explaining \ac{ML} models produced by \ac{AutoML}. In a requirement analysis study, we have collected desired explanations and information from a diverse user basis containing 16 data scientists, 11 domain experts, and 9 \ac{AutoML} researchers. The results indicate that designing visual analytics tools focused on a single user group is, in the context of \ac{AutoML}, not helpful. A clear assignment of explanations to specific user groups is often not possible. \name{XAutoML} allows users to pick interesting information from a broad spectrum. In a user study, we were able to prove the effectiveness and usefulness of \name{XAutoML}. Our results showed that the usage of \name{XAutoML} significantly increased the understanding of \ac{AutoML} systems and \ac{ML} models produced by them leading to an increased trust of users in \ac{AutoML}.

\begin{acks}

The authors wish to thank the participants of the requirements analysis and the user study for their time and valuable input. This work was partially supported by the German Federal Ministry for Economic Affairs and Climate Action in the project FabOS (project no. 01MK20010N), by the German Federal Ministry of Education and Research in the project RESPOND (project no. 01|S8067A) and by the Baden-Wuerttemberg Ministry for Economic Affairs, Labour
and Tourism in the project KI-Fortschrittszentrum ``Lernende Systeme und Kognitive Robotik'' (project no. 036-170017).

\end{acks}

\bibliographystyle{ACM-Reference-Format}
\bibliography{library}

\appendix

\section{Detailed Study Results}

\begin{table}[hbt]
    \centering
    
    \caption{Summary of the questionnaire measuring the prior knowledge of domain experts (DE), data scientists (DS), \ac{AutoML} researcher (AR), and all together in \ac{AutoML} related topics. Values are given on a scale of 1 (strongly disagree) to 5 (strongly agree).}
    \label{tab:prior_knowledge}
    
    \begin{tabularx}{\textwidth}{l X c c c c}
        \toprule
        &                                                           & DE                & DS                & AR                & All \\
        \midrule
        P01 & I am confident in building my own ML pipelines.       & \(2.32 \pm 1.34\)	& \(3.63 \pm 0.97\)	& \(3.70 \pm 0.67\)	& \(3.00 \pm 1.39\) \\
        P02 & I am familiar with AutoML.                            & \(1.95 \pm 1.03\)	& \(2.89 \pm 1.50\)	& \(4.20 \pm 1.48\)	& \(2.56 \pm 1.46\) \\
        P03 & I believe AutoML produces good models.                & \(2.63 \pm 0.96\)	& \(3.30 \pm 0.82\)	& \(3.40 \pm 0.70\)	& \(3.03 \pm 1.00\) \\
        \bottomrule
    \end{tabularx}
\end{table}

\begin{table}[hbt]
    \centering
    
    \caption{Table of all cards containing potential information presented during the card-sorting requirements analysis. Given is the mean rank of each card for domain experts (DE), data scientists (DS), \ac{AutoML} researchers (AR), and all. A rank of \(1\) equals the most important information, a rank of \(24\) the least important. The complete list of all cards, including the example visualizations and textual explanations, is available at \url{https://github.com/Ennosigaeon/xautoml/blob/master/user_study/card_sorting_examples.pdf}.}
    \label{tab:card_sorting_results}

    \begin{tabularx}{\textwidth}{l X c c c c}
        \toprule
               &                                                                                          & DE          & DS        & AR        & All \\
        \midrule
        R01    & View the input data                                                                      & \(13.1\)	& \(19.4\)	& \(14.4\)	& \(15.8\) \\
        R02    & Know context information about features in the input data                                & \(7.9\)	    & \(16.4\)	& \(19.0\)	& \(13.9\) \\
        R03    & View statistics of input data                                                            & \(10.3\)	& \(18.4\)	& \(14.1\)	& \(14.3\) \\
        R04    & Visualize the input data                                                                 & \(11.3\)	& \(16.8\)	& \(10.7\)	& \(13.2\) \\
        R05    & View the pre-processed data                                                              & \(13.7\)	& \(17.5\)	& \(19.2\)	& \(16.5\) \\
        R06    & Know how input data was pre-processed                                                    & \(15.4\)	& \(11.8\)	& \(13.0\)	& \(13.5\) \\
        R07    & View statistics of pre-processed data                                                    & \(10.0\)	& \(12.2\)	& \(16.4\)	& \(12.5\) \\
        R08    & Visualize pre-processed data                                                             & \(13.2\)	& \(12.8\)	& \(18.1\)	& \(14.3\) \\
        R09    & View data after features engineering                                                     & \(9.2\)	    & \(12.5\)	& \(15.6\)	& \(12.1\) \\
        R10    & View statistics of data after feature-engineering                                        & \(11.2\)	& \(13.3\)	& \(17.7\)	& \(13.6\) \\
        R11    & Visualize data after feature-engineering                                                 & \(10.0\)	& \(12.0\)	& \(16.3\)	& \(12.4\) \\
        R12    & Know how new features are derived from existing features                                 & \(11.8\)	& \(10.4\)	& \(14.6\)	& \(12.0\) \\
        R13    & View the complete processing pipeline                                                    & \(16.2\)	& \(8.5\)	& \(5.9\)	& \(10.7\) \\
        R14    & Know which pre-processing, feature engineering, and modeling algorithms are available    & \(15.8\)	& \(10.3\)	& \(8.9\)	& \(12.0\) \\
        R15    & Know how pipelines are chosen                                                            & \(17.5\)	& \(10.2\)	& \(6.4\)	& \(11.9\) \\
        R16    & Know how hyperparameters are chosen                                                      & \(17.8\)	& \(14.7\)	& \(7.7\)	& \(14.0\) \\
        R17    & View performance metrics                                                                 & \(2.8\)	    & \(4.8\)	& \(6.7\)	& \(4.5\) \\
        R18    & Visualize evaluation metrics                                                             & \(9.4\)	    & \(8.2\)	& \(8.0\)	& \(8.6\) \\
        R19    & View global surrogates for model                                                         & \(11.5\)	& \(10.6\)	& \(11.9\)	& \(11.3\) \\
        R20    & View local surrogates for model                                                          & \(9.5\)	    & \(9.8\)	& \(13.7\)	& \(10.7\) \\
        R21    & View hyperparameters of model                                                            & \(17.7\)	& \(13.8\)	& \(9.4\)	& \(14.1\) \\
        R22    & Compare performance and explanations of various models with each other                   & \(10.6\)	& \(9.7\)	& \(9.6\)	& \(10.0\) \\
        R23    & Compare pipeline structures with each other                                              & \(17.5\)	& \(10.8\)	& \(10.9\)	& \(13.3\) \\
        R24    & Compare hyperparameters of identical pipelines                                           & \(16.8\)	& \(15.2\)	& \(11.9\)	& \(14.9\) \\
        \bottomrule
    \end{tabularx}
\end{table}

\begin{table}
    \centering
    \caption{Set of the tasks given to the study participants. Tasks marked by \textit{All} are performed by all participants, tasks marked by \textit{Experts} only by participants who are proficient in the according dimension.}
    \label{tab:study_tasks}

    \begin{tabularx}{\textwidth}{l l X c}
        \toprule
            &           &                                                                       & Correct \% \\
        \midrule
        \multicolumn{3}{l}{Optimization Overview} \\
        T01 & All       & What performance does the best pipeline have?                         & 100 \\
        T02 & All       & Is the best pipeline performing significantly better than the rest?   & 100 \\
        T03 & All       & Are better pipelines found over time?                                 & 94 \\
        T04 & All       & How many pipelines were generated in total?                           & 100 \\
        T05 & All       & How long did the optimization take?                                   & 100 \\
        T06 & All       & How was the optimization configured?                                  & 94 \\  
        \\
        \multicolumn{3}{l}{Domain Expertise} \\
        T10 & All       & What performance does the model have regarding specific metrics?      & 100 \\
        T11 & All       & Why was a single sample predicted as <CLASS>?                         & 94 \\
        T12 & All       & How does this model approximately work?                               & 100 \\
        T13 & All       & Which features are important?                                         & 97 \\
        T14 & Experts   & Does the model produce sensible predictions?                          & 95 \\
        T15 & Experts   & Does the model select reasonable features for the predictions?        & 100 \\
        T16 & Experts   & Are there superfluous features in the dataset?                        & 100 \\
        \\
        \multicolumn{3}{l}{\ac{ML} Expertise} \\
        T20 & All       & What steps does the pipeline have?                                    & 97 \\
        T21 & All       & Which classifier does the pipeline use?                               & 100 \\
        T22 & All       & How fast is a prediction using this pipeline?                         & 97 \\
        T23 & Experts   & Which are the hyperparameters of this pipeline?                       & 93 \\
        T24 & Experts   & Which are the most important hyperparameters?                         & 96 \\
        T25 & Experts   & What range should <HYPERPARAMETER> be to yield a good performance?    & 85 \\
        \\
        \multicolumn{3}{l}{\ac{AutoML} Expertise} \\
        T30 & All       & Can you explain how the search algorithm for pipelines works?         & 92 \\
        T31 & All       & Can you explain how the search algorithm for hyperparameters works?   & 83 \\
        T32 & All       & How many hyperparameters does the random forest have?                 & 97 \\
        T33 & Experts   & What does the search space look like?                                 & 100 \\
        T34 & Experts   & Does the search algorithm converge to a local minimum?                & 100 \\
        T35 & Experts   & Do you think the underlying search space is well selected?            & 89 \\
        \bottomrule
    \end{tabularx}
    
\end{table}

\begin{table}
    \centering
    \caption{Detailed results of the \ac{SUS} questionnaire for domain experts (DE), data scientists (DS), \ac{AutoML} researchers (AR), and all.}
    \label{tab:sus}

    \begin{tabularx}{\textwidth}{X l l l l}
        \toprule
                                                                                            & DE   & DS   & AR   & All \\
        \midrule
        I think that I would like to use this system frequently.                            & 7.50 & 7.98 & 7.50 & 7.64 \\
        I found the system unnecessarily complex.                                           & 7.63 & 8.08 & 8.33 & 7.93 \\
        I found the system was easy to use.                                                 & 7.50 & 7.12 & 7.50 & 7.21 \\
        I think that I would need support of a technical person to be able to use this system. & 7.89 & 8.37 & 8.33 & 8.00 \\
        I found the various functions in this system were well integrated.                  & 7.63 & 8.17 & 8.06 & 7.86 \\
        I thought there was too much inconsistency in this system.                          & 8.42 & 8.75 & 8.61 & 8.57 \\
        I would imagine that most people would learn to use this system very quickly.       & 7.11 & 7.40 & 7.22 & 7.14 \\
        I found the system very cumbersome to use.                                          & 7.89 & 8.27 & 7.50 & 8.14 \\
        I felt very confident using the system.                                             & 7.37 & 7.60 & 7.22 & 7.36 \\
        I needed to learn a lot of things before I could get going with this system.        & 7.37 & 8.08 & 8.61 & 7.65 \\
        \bottomrule
    \end{tabularx}
\end{table}

\end{document}